\documentclass[runningheads]{llncs}

\usepackage[T1]{fontenc}
\usepackage{graphicx}
\usepackage{booktabs}
\usepackage{multirow}
\newcommand{\corr}{(*)}

\usepackage{amsmath}
\usepackage{subcaption}
\usepackage{nicefrac}
\usepackage{hyperref}
\usepackage{orcidlink}
\usepackage{booktabs}
\usepackage{dsfont}
\usepackage{tikz}
\usetikzlibrary{arrows}
\usepackage{blkarray}
\usetikzlibrary{positioning}
\usetikzlibrary{fit}
\usepackage{placeins}
\usepackage{xspace}

\newcommand{\indic}{\mathds{1}}
\newcommand{\E}{\mathds{E}}
\newcommand\Proba{{\mathds{P}}}

\newcommand{\ones}{\boldsymbol{1}}

\newcommand\lca{\breve{a}}

\newcommand\loss{L}

\newcommand\bbeta{{\boldsymbol{\beta}}}

\newcommand\btheta{{\boldsymbol{\theta}}}
\newcommand\bx{{\mathbf{x}}}
\newcommand\bz{{\mathbf{z}}}
\newcommand\clE{{\mathcal{E}}}

\newcommand\clV{{\mathcal{V}}}
\newcommand\clW{{\mathcal{W}}}
\newcommand\clX{{\mathcal{X}}}

\newcommand\rmf{{\mathrm{f}}}
\newcommand\p[1]{{\left(#1\right)}}

\newcommand{\abs}[1]{\left|#1\right|}

\definecolor{mygreen}{rgb}{0, 0.75, 0}
\definecolor{mixredg}{rgb}{1, 0.5, 0}
\definecolor{mixredw}{rgb}{1, 0.25, 0.25}
\definecolor{mixredb}{rgb}{1, 0, 0.5}
\definecolor{mixbluer}{rgb}{0.5,0,1}
\definecolor{mixblueg}{rgb}{0,0.5,1}
\definecolor{mixbluegr}{rgb}{0.25,0.5,1}
\definecolor{mixbluegg}{rgb}{0,0.75,1}

\definecolor{ours}{RGB}{44,160,44}
\definecolor{bertinetto}{RGB}{255, 127, 14}
\definecolor{crossentropy}{RGB}{31, 119, 180}

\definecolor{bertinetto1}{RGB}{220,20,60}
\definecolor{bertinetto5}{RGB}{139,0,0}
\definecolor{crossentropy}{RGB}{135,206,235}
\definecolor{hier9}{RGB}{127,255,0}
\definecolor{hier12}{RGB}{50,205,50}

\makeatletter
\DeclareRobustCommand\onedot{\futurelet\@let@token\@onedot}
\def\@onedot{\ifx\@let@token.\else.\null\fi\xspace}

\def\vs{\emph{vs}\onedot}

\def\etal{\emph{et al}\onedot}
\makeatother

\tikzset{
  treenode/.style = {align=center, inner sep=0pt, text centered, font=\sffamily},
  root/.style = {treenode, circle, black, font=\sffamily\bfseries, draw=black, fill=white, text width=1.5em},
  node/.style = {treenode, circle, white, font=\sffamily\bfseries, text width=1.5em},
  leaf/.style = {treenode, rectangle, white, font=\sffamily\bfseries, text width=1.5em, minimum width=1.5em, minimum height=1.5em},
  empty/.style = {treenode}
}

\newcommand\textpropositionweighting[1]{%
  The exponential weighting scheme of Equation~\eqref{#1} produces a balanced
  weighted tree, where the weights along the path from the root to any
  leaf sum up to \nicefrac{1}{2}, following an exponential
  growth/decay according to the value of parameter $q$, such that
  $w_j=q \, w_{p\p{j}}$ for all children of node $v_{p(j)}$ that are
  at height $h(p(j))-1$.%
}

\newcommand\textpropositionloss[1]{%
  The loss function~\eqref{#1} is a proper scoring
  rule,
  meaning that the true conditional probabilities $\Proba(Y=k\mid
  X=\bx)$ minimize the expectation of
  the loss with respect to $(Y\mid X=\bx)$,
  if and only if the weighting scheme of the tree is balanced.%
}

\newcommand\textpropositionnotproperloss[1]{%
  If the number of ancestors of each leaf is not identical, then the
  loss function~\eqref{#1} is not a proper
  scoring rule, meaning that the true conditional probabilities
  $\Proba(Y=k\mid X=\bx)$ do not minimize the expectation of the loss
  with respect to $(Y\mid X=\bx)$.
}

\newif\ifverylong
\verylongfalse

\newif\ifdraft
\drafttrue

\newcommand\ycomment[1]{{}}
\newcommand\lcomment[1]{{}}
\newcommand\solved[1]{{}}

\makeatletter
\DeclareRobustCommand\onedot{\futurelet\@let@token\@onedot}
\def\@onedot{\ifx\@let@token.\else.\null\fi\xspace}

 \def\vs{\emph{vs}\onedot}

\def\etal{\emph{et al}\onedot}
\makeatother

\begin{document}

\title{Harnessing Superclasses for Learning from Hierarchical Databases} 

\author{Nicolas Urbani\inst{1} \corr \and
  Sylvain Rousseau\inst{1}\and
  Yves Grandvalet\inst{1}\and
  Leonardo Tanzi\inst{2} }

\toctitle{Harnessing Superclasses for Learning from Hierarchical Databases}
\tocauthor{Nicolas Urbani, Sylvain Rousseau, Yves Grandvalet,Leonardo Tanzi}

\authorrunning{N. Urbani et al.}

\institute{Université de technologie de Compiègne, CNRS, Heudiasyc UMR 7253, France \and Politecnico di Torino, Italy}
\maketitle              %

\begin{abstract}
  In many large-scale classification problems, classes are organized in a
  known hierarchy, typically represented as a tree expressing the inclusion of
  classes in superclasses. We introduce a loss for this type of
  supervised hierarchical classification. It utilizes the knowledge of the
  hierarchy to assign each example not only to a class but also to all
  encompassing superclasses. Applicable to any feedforward architecture with a
  softmax output layer, this loss is a proper scoring rule, in
  that its expectation is minimized by the true posterior class probabilities.
  This property allows
  us to simultaneously pursue consistent classification objectives between
  superclasses and fine-grained classes, and eliminates the need for a
  performance trade-off between different granularities. We conduct an
  experimental study on three reference benchmarks, in which we vary the size of
  the training sets to cover a diverse set of learning scenarios. Our approach
  does not entail any significant additional computational cost compared with
  the loss of cross-entropy. It improves accuracy and reduces the number of
  coarse errors, with predicted labels that are distant from ground-truth labels
  in the tree.

\end{abstract}

\section{Introduction}
\label{sec:intro}

This paper proposes a family of training losses for hierarchical
supervised classification (HSC), when classes are organized in a tree
structure representing the relationships between
classes~\cite{blockeel_2002}.  
In this framework, classes are the leaves of the tree, and can be
grouped into coarser superclasses until the root node is reached.  The
knowledge of superclasses is an additional piece of information
compared to the simple fine-grained class information, and can
therefore improve training efficiency.
The hierarchy can also be used to define and reduce the severity of the errors; for example, a confusion between ``cat'' and ``dog'' may be deemed less serious than a confusion between ``cat'' and ``airplane''.

The majority of existing
approaches for HSC~\cite{3_zhu_2017,tanzi_hierarchical,4_koo_2020,10_yan_2015,6_wang_2021}
rely on ad-hoc architectures to exploit the class hierarchy, for example by
adding intermediate outputs~\cite{3_zhu_2017}, or on cascades of separately
trained classifiers~\cite{tanzi_hierarchical}. In contrast, our approach 
redefines the objectives of learning, regardless of the model, by defining a
training loss that can be applied to any architecture with a final softmax
layer. 

We test this approach on the classical benchmarks
iNaturalist19~\cite{van2018inaturalist},
TinyImageNet~\cite{chrabaszcz2017downsampled}, and a modified version of
ImageNet-1K~\cite{imagenet}, in which we vary the number of samples used for
training in order to test the effectiveness of our approach in a variety of
learning scenarios. These experiments are conducted with two well-known
architectures: ResNet50~\cite{he2016deep} and
MobileNetV3-Small~\cite{howard2019searching}, two convolutional networks that
differ notably in their number of parameters.

Our contributions are threefold: 
\begin{itemize}
    \item we propose a large family of training losses that are proper scoring rules;
    \item we show that previous work that was motivated by very different arguments also has this property;
    \item we propose two new measures of performance that take into account the hierarchy, one based on optimal transport, the other one being a curve summarizing the performances of a model used for solving the hierarchical classification problem at all granularities.
\end{itemize}

This paper is structured as follows. Section~\ref{sec:2} provides an overview of
related work. Section~\ref{sec:loss} details the design and properties of the proposed
loss, and Section~\ref{sec:4} presents the experimental protocol followed to
obtain the results presented, discussed and analyzed in Section~\ref{sec:5}. We
conclude in Section~\ref{sec:6} which also gives some suggestions for future work. The source code is available on the following repository: \href{https://github.com/heudiasyc/Harnessing-Superclasses-for-Learning-from-Hierarchical-Databases}{https://github.com/heudiasyc/Harnessing-Superclasses-for-Learning-from-Hierarchical-Databases}.

\section{Related Works} \label{sec:2}

Hierarchical supervised classification (HSC) has received substantial attention
in computer vision and beyond. In contrast to standard supervised
classification, HSC attempts to classify the objects into a hierarchy of
categories, with each level of the hierarchy corresponding to a level of
conceptual abstraction. 

The standard multiclass approach ignores the class
hierarchy and fits a ``flat'' classifier, which implicitly assumes that all
classes are equidistant. As HSC labels are richer, they can help improve
performance when class similarity in the hierarchy matches visual similarity.

\paragraph{HSC tasks.} Label hierarchies can be used for several purposes. 
In large-scale applications, they can play a decisive role in the design of specific approaches for processing massive amounts of data~\cite{1_babbar_2016,8_zhao_2011,9_goo_2016,11_wu_2017}. 
They can also be used for improving accuracy by allowing to predict superclasses instead of classes for ambiguous items~\cite{15_wu_2020}.
When no class structure is known, possible hierarchies can be
inferred~\cite{10_yan_2015,12_srivastava_2013,13_salakhutdinov_2011,nister2006scalable,15_wu_2020}.

\paragraph{Pipelines for HSC.} Some methods modify the processing pipeline to
handle the hierarchy.
A first line of approach makes use of the specific characteristics of the
representations of examples at different layers of CNNs, with the
representations going from low-level to high-level and from local to global.
These representations at different layers can be used to classify examples at
different levels of the hierarchy \cite{3_zhu_2017,Chang_2021_CVPR,garg_eccv_2022}.
A
second strategy is to use a cascade of models to classify into superclasses
following the hierarchy (e.g.,\cite{tanzi_hierarchical}), but errors accumulate
along the cascade and the last models see only a very small part of the data, which is particularly problematic on small data sets. Finally, the
hierarchy can be taken into account by the post-processing of standard ``flat''
classifiers,
for example by using conditional random fields \cite{7_deng_2014}, or with post-hoc approaches \cite{karthik2021no}.

\paragraph{Losses for HSC.} Several training losses have been proposed to handle
hierarchical labels. 
Wang \etal \cite{6_wang_2021}, Wu \etal \cite{wu_hierarchical_2019} and Wu \etal \cite{15_wu_2020} define specific losses that
can be applied to any existing architecture, but their objectives differ from ours.
Chang \etal \cite{Chang_2021_CVPR} use a convex combination of
cross-entropies at different levels of granularity, 
without any connection between class and superclass scores beyond their common representation space.
Bertinetto \etal \cite{bertinetto2020making} propose two training methods, based on soft labels or on the chaining of conditional probabilities of child classes knowing their parent. 
The latter method is the closest relative to the present proposal, and will be detailed in Section~\ref{sec:bertinetto}.

Here, we deal exclusively with known hierarchies, as in ImageNet or iNaturalist. 
There are several hierarchies of this type
for animals, plants, fungi or artifacts. We also focus on situations where data
is scarce, with the aim to
improve classification in two ways: by reducing the error rate, and by avoiding
the coarsest errors, between labels that are farthest apart in the hierarchy.
Although our framework allows it, we do not consider partial
classification at non-terminal nodes of the
hierarchy~\cite{15_wu_2020}, but we provide experimental results that
illustrate how our proposal improves over the standard learning
protocol based on cross-entropy when allowing to aggregate
fine-grained classes into coarser superclasses.

Our proposal differs from all pipelines approaches
in that we do not modify the network structure, we do not add outputs, and we do
not post-process these outputs. 
Learning differs from standard
learning only in the training loss that uses the hierarchy to estimate
fine-grained classes, and where superclass scores are
consistent with fine-grained class scores, so as to help structure the predictions.

Instead of establishing a trade-off between antagonistic objectives at different levels of
granularity, we use weights to place more or less emphasis on the
recognition on some superclasses, according to their position in the
hierarchy.  Our aim is to produce a solution that provides a
satisfactory answer to all classification problems that can be defined
from any partition of classes that is consistent with the hierarchy.
We only assign posterior probabilities to
fine-grained classes, which ensures consistent scorings, in the sense
that all probabilities assigned to superclasses are guaranteed to be
the sum of the probabilities of their child classes.

\section{Hierarchical Loss}\label{sec:loss}

We introduce here the notation describing the class hierarchy. We then describe
a family of weighting schemes that can be used to define distances between
classes in this hierarchy. Finally, we define and motivate the training loss
designed to account for this hierarchy.
All proofs are given in Appendix \ref{sec:proofs}

\subsection{Notation}

Let $\mathcal{D}=\{\bx_{i},y_{i}\}_{i=1}^n$ be the training set, where
$\bx_{i} \in \clX$ is an input and $y_{i} \in \{1,\ldots,K\}$ is its label, and
$\btheta$ be the parameters of the model to be trained. The model output
corresponding to class $k$ is denoted $f_{k}(\bx_{i};\btheta)$. It is obtained
by a softmax layer applied to the representation of the input $\bx_{i}$ on the
penultimate layer, denoted $\bz_{i}$ ($\bz_{i} \in \mathds{R}^{m}$, where $m$ is
the dimension of the penultimate layer):
\begin{equation}\label{eq:softmax}
  f_{k}(\bx_{i};\btheta) = \frac{\exp(\bbeta_k^T\bz_{i}+\alpha_k)}{\sum_{l=1}^K\exp(\bbeta_l^T\bz_{i}+\alpha_l)}
  \enspace.
\end{equation}
The parameters $\alpha_k\in \mathds{R}$ and  
$\bbeta_k\in \mathds{R}^{m}$ correspond to the (fine-grained) class $k$. Without
subscripts, $(\bx,y,\bz)$ represent a generic input, output and representation
respectively, and $(X,Y,Z)$ are the corresponding random variables.

The class hierarchy is represented by a weighted directed rooted tree
$T=(\clV,\clE,\clW)$ whose nodes $v_{j} \in \clV$ represent subsets of classes,
and where an edge $ e_{jj'} = (v_{j},v_{j'}) \in \clE$ indicates that the subset
of labels $j$ subsumes the labels $j'$, that is, $j$ is a superclass of $j'$.
The tree has thus $K$ leaves representing the $K$ classes $\{y_k\}$, and the
root represents the set of all classes $\{y_1,\ldots,y_k,\ldots,y_K\}$. The
weights $w_{j}\in\clW$ are attached to $v_{j}$, or equivalently to the incoming
edge of $v_{j}$; our weighting scheme is described in
Section~\ref{sec:weighting}. Let $c({j})$ be the set of indices of the children
of $v_{j}$, 
and $p({j})$ be the index of the parent node of $v_{j}$.
The set of indices of all its ancestors is denoted $a({j})$, where, to match our needs,
the root is excluded from $a({j})$ and ${j}$ is included in $a({j})$. In words,
the ancestors of node $v_{j}$ are the nodes on the path from $v_{j}$ to the
root, root excluded. 
The index of the lowest common ancestor of  $v_{j}$ and  $v_{j'}$ is denoted $\lca(j,j')$.
Finally, let $h(j)$ be the height of $v_{j}$: the number of
edges from the node to the deepest leaf. All leaves have zero height and each
non-leaf node $v_j$ has height $1+\max_{j'\in{c(j)}} h(j')$.

We use the convention that the root node is indexed by $0$, and that indexing of
nodes then starts from the leaves, so that, in a classification problem with $K$
classes, for all $k$ such that $0<k\leq K$, $v_{k}$ is a singleton leaf node
representing the fine-grained class $k$.
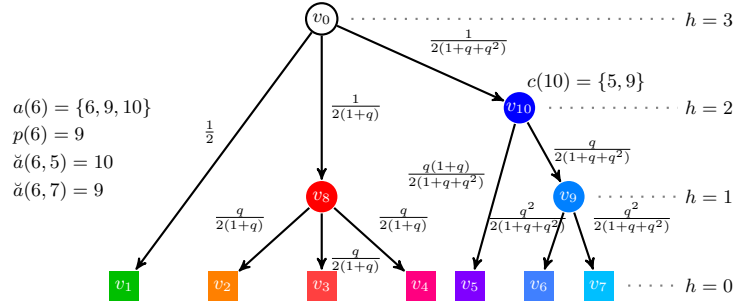
\begin{figure}[t]
    \begin{center}%
    \resizebox{0.8\textwidth}{!}{%
    \begin{tikzpicture}[->,>=stealth', line width = 1pt,
    level 1/.style={level distance=4.5em,sibling distance=10em},
    level 2/.style={level distance=4.5em,sibling distance=5em},
    level 3/.style={level distance=4.5em,sibling distance=3em}
    ] 
    \node(Root) [root] {$v_{0}$}
    child[level distance=13.5em]{ node [leaf,fill=mygreen] {$v_{1}$} edge from parent node[above left] {$\frac{1}{2}$}}
    child[level distance=9em]{ node [node,fill=red] {$v_{8}$} 
      child{ node [leaf,fill=mixredg] {$v_{2}$} edge from parent node[above left]  {$\frac{q}{2(1+q)}$}}
      child{ node [leaf,fill=mixredw] {$v_{3}$} edge from parent node[below right] {$\frac{q}{2(1+q)}$}}
      child{ node [leaf,fill=mixredb] {$v_{4}$} edge from parent node[above right] {$\frac{q}{2(1+q)}$}}
    edge from parent node[right] {$\frac{1}{2(1+q)}$}}
    child{ node [node,fill=blue] {$v_{10}$} 
      child[level distance=9em]{ node [leaf,fill=mixbluer] {$v_{5}$}  edge from parent node[above left] {$\frac{q(1+q)}{2(1+q+q^2)}$}}
      child{ node [node,fill=mixblueg] {$v_{9}$} 
        child{ node [leaf,fill=mixbluegr] {$v_{6}$} edge from parent node[above]  {$\frac{q^2}{2(1+q+q^2)}$~~~~~~~~~}}
        child{ node [leaf,fill=mixbluegg] {$v_{7}$} edge from parent node[above right] {$\frac{q^2}{2(1+q+q^2)}$}}
      edge from parent node[right] {$\frac{q}{2(1+q+q^2)}$}}
    edge from parent node[above right] {$\frac{1}{2(1+q+q^2)}$}}
  ; 
   \begin{scope}[every node/.style={right}]
     \path (Root       -| Root-3-2-2) ++(4em,0) node(d0) {$h=3$};
     \path (Root-3     -| Root-3-2-2) ++(4em,0) node(d1) {$h=2$};
     \path (Root-3-2   -| Root-3-2-2) ++(4em,0) node(d2) {$h=1$};
     \path (Root-3-2-2 -| Root-3-2-2) ++(4em,0) node(d3) {$h=0$};
     \path (Root -| Root-3) ++(-0em,-3.2em) node {$c({10})=\{{5},{9}\}$};
   \end{scope}
  \path[draw, loosely dotted,gray,-] (Root) ++(1.5em,0) edge (d0);
  \path[draw, loosely dotted,gray,-] (Root-3) ++(1.5em,0) edge (d1);
  \path[draw, loosely dotted,gray,-] (Root-3-2) ++(1.5em,0) edge (d2);
  \path[draw, loosely dotted,gray,-] (Root-3-2-2) ++(1.5em,0) edge (d3);
   \begin{scope}[every node/.style={right}]
     \path (Root-3 -| Root-1) ++(-6em,0em)  node {$a({6})=\{{6},{9},{10}\}$};
     \path (Root-3 -| Root-1) ++(-6em,-1.4em)  node {$p({6})={9}$};
     \path (Root-3 -| Root-1) ++(-6em,-2.8em)  node {$\lca(6,5)=10$};
     \path (Root-3 -| Root-1) ++(-6em,-4.2em)  node {$\lca(6,7)=9$};
   \end{scope}
   \end{tikzpicture}}
   \caption{Illustration of a tree structure recalling the notation and weighting system}
   \label{fig:treedecomposition}
  \end{center}
\end{figure}
Figure~\ref{fig:treedecomposition} shows an example of a tree that illustrates
these definitions and conventions, together with a balanced weighting defined in
the next section. In this example, the singleton leaves are
$\{v_{1}, \ldots, v_{7}\}$, and the superclasses are
$v_{8}=\{v_{2}, v_{3}, v_{4}\}$, $v_{9}=\{v_{6}, v_{7}\}$ and
$v_{10}=\{v_{5}, v_{9}\}$.

\subsection{Weighting Scheme}\label{sec:weighting}

\paragraph{Balanced Weighted Tree.} Tree weighting plays a key role to ensure
that our hierarchical loss is a proper scoring rule~\cite{gneiting2007strictly}.
Here, we will consider several choices, all of which have in common that the
weight attached to the nodes is zero at the root, and strictly positive
elsewhere. Furthermore, the weighting forms a balanced weighted tree, meaning
that the sum of the weights on the path from any leaf to the root is constant.
Without loss of generality, this constant is set to~\nicefrac{1}{2}, so as to
relate distance to (discounted hierarchical) errors.

The distance between any leaf and the root is thus \nicefrac{1}{2}. If two
leaves $\{\hat{y}\}$ and $\{y\}$ have no common ancestor but the root node, the
distance between $\{\hat{y}\}$ and $\{y\}$ is 1; if node $v_{j}$ is both on the
path from $\{\hat{y}\}$ to the root and on the path from $\{y\}$ to the root,
then $v_{j}$ is a common ancestor of $\{\hat{y}\}$ and $\{y\}$, and this
proximity between $\{\hat{y}\}$ and $\{y\}$ leads to a reduction of distance of
$w_j$. Overall the distance between $\{\hat{y}\}$ and $\{y\}$ in the weighted
tree $T$ is:
\begin{align}
  d_{T}(\hat{y},y) & = \Big( \sum_{j\in a({\hat{y}})} w_j - \! \!   \sum_{j\in \lca({y},\hat{y})} \! \!  w_j \Big) 
                     + \Big(\sum_{j\in a({y})} w_j - \! \!   \sum_{j\in \lca({y},\hat{y})} \! \!   w_j \Big) \label{eq:d_T} \\
                   & = 1- 2 \sum_{j\in \lca({y},\hat{y})} w_j \nonumber \\
                   & = 2 \sum_{j\in a({\hat{y}}) \setminus \lca({y},\hat{y})} w_j \enspace = 2 \sum_{j\in a({y}) \setminus \lca({y},\hat{y})} w_j
                     \enspace. \nonumber
\end{align}
The first equation states the usual definition of a distance between nodes: the
shortest path from $\{\hat{y}\}$ to $\{y\}$ is the sum of the distances from (i)
$\{\hat{y}\}$ to the lowest common ancestor of $\{\hat{y}\}$ and and $\{y\}$,
and from (ii) $y$ to the lowest common ancestor of $\{\hat{y}\}$ and $\{y\}$.
The other equations are properties derived from the balancing of the weighted
tree. The second equation states that this distance can also be calculated
solely from the path from the lowest common ancestor to the root, and the
last ones that it can also be computed from any path between one of the two
leaves and their lowest common ancestor.

\paragraph{Exponential Weighting.}

The weighting is parameterized by $q \geq 0$, and is recursively defined as follows:
\begin{equation}
  \label{eq:2}
  \forall j\neq0,\enspace w_j = \Big( \frac{1}{2}-\sum_{k\in a(p(j))} w_{k} \Big) \frac{1-q}{1-q^{h\p{j}+1} }
  \enspace,
\end{equation}
with $w_0=0$ for the root.

\begin{proposition}\label{prop:1}
  \textpropositionweighting{eq:2}
\end{proposition}

\begin{remark}
  If all non-leaf nodes have their children at the same height, then
  $w_j=q\, w_{p\p{j}}$ for all non-root nodes. This holds in particular for
  perfect trees.
\end{remark}

Depending on the value of $q$, the weighting scheme gives more or less
importance to superclasses according to their height in the tree. This is
illustrated in Figure~\ref{fig:treedecomposition} with the generic expression of
the weights. For $q=1$, all weights are roughly equal. For $q<1$, the most
important weights are the ones close to the root, and for $q=0$, the distance
between classes is only determined by their membership of the largest
superclasses: there is no more differentiation of classes within these
superclasses, which are all equidistant. For $q>1$ the most important weights
are at the leaves of the tree, and as $q$ goes to infinity, the distance between
classes is no more determined by the hierarchy, since all classes are
equidistant, as in standard multi-class classification.
\subsection{Hierarchical Loss}
\label{sec:hl}

The class hierarchy is taken into account during learning by using a loss
function that rewards all partial correct answers given by the model: when
$y=k$, all the superclasses that include class $k$ are partially correct. The
simplest option would be to reward all correct partial answers given at each
node of the class hierarchy, so that the ancestors of leaf $k$ in the hierarchy
could be rewarded to provide that partial answer. Such a loss would read as
follows:
\begin{equation}\label{eq:hierarchicalloss_init}
  - \sum_{v_{j}\in a({y})} \log\Big(\sum_{k\in v_{j}} f_{k}(\bx;\btheta)\Big)
  \enspace.
\end{equation}
With the standard cross-entropy loss, $f_{k}(\bx;\btheta)$ estimates
$\Proba(Y=k\mid X=\bx)$, in which case $\sum_{k\in v_{j}} f_{k}(\bx;\btheta)$
estimates $\Proba(Y\in v_{j}\mid X=\bx) = \sum_{k\in v_{j}} \Proba(Y=k\mid
X=\bx)$.
However, the following proposition indicates that, in general, the loss~\eqref{eq:hierarchicalloss_init} is not appropriate for training.

\begin{proposition}\label{prop:not_proper_loss}
  \textpropositionnotproperloss{eq:hierarchicalloss_init}
\end{proposition}

Intuitively, one understands that this loss is biased by considering
that the number of ancestors of classes in the hierarchy has an impact on the
number of occurrences of examples of these classes in the loss.
Instead, we propose a simple loss, using the weighted tree representing the
class hierarchy, as follows:
\begin{equation}\label{eq:hierarchicalloss}
  \loss_{T}(f(\bx;\btheta),y) =
  - \sum_{j\in a(y)} w_{j} \log\Big( \sum_{k\in v_{j}} f_{k}(\bx;\btheta)\Big)
  \enspace.
\end{equation}
For a ``flat hierarchy'', where each leaf is directly attached to the root, $y$
is its only ancestor ($a(v_{y})=\{y\}$) and there is only one element in the
subset $v_{y}=\{y\}$, so that we recover the standard cross-entropy loss
multiplied by \nicefrac{1}{2} (note that this is the limiting case of
$q\rightarrow +\infty$ in our exponential weighting scheme).

\begin{proposition}\label{prop:proper_loss}
	\textpropositionloss{eq:hierarchicalloss}
\end{proposition}

Our approach to computing the loss is based on matrix computations that
make the procedure fast and adaptable to any tree, so that the code can handle
any class hierarchy. This additional processing
corresponds to a very sparse, non-trainable linear layer and has no significant effect on the computational cost of a training epoch.

\subsection{Comparison with Hierarchical Cross-Entropy}\label{sec:bertinetto}

The hierarchical cross-entropy (HXE) \cite{bertinetto2020making} relies on the hierarchy to decompose the posterior probabilities along the path from the root to leaf $k$ :
\begin{equation}\label{eq:bertinetto:decomposition}
	\Proba(Y=k\mid X=\bx) = \prod_{j\in a(k)} \Proba(Y\in v_{j}\mid
	Y\in v_{p(j)}, X=\bx) 
	\enspace, 
\end{equation}
from which they propose the following loss:
\begin{equation}\label{eq:bertinetto}
	\loss_{\text{HXE}}(f(\bx;\btheta),y) =
	 -\sum_{j\in{a\p{y}}} %
	 \exp\p{-\alpha d\p{j}}
  \log\p{\frac{\sum_{k\in v_{j}} f_{k}(\bx;\btheta)}{\sum_{k\in v_{p(j)}}
	 f_{k}(\bx;\btheta)}}
\end{equation}
where $d\p{j}$ is the depth of node $v_{j}$ and $\alpha>0$ is a
trade-off parameter that controls whether the focus is on fine-grained
classes (small $\alpha$ values) or on large superclasses (large
$\alpha$ values).

This derivation is reminiscent of a soft cascade of classifiers, since the \eqref{eq:bertinetto:decomposition} decomposition is a probabilistic way of concatenating classifiers operating at increasingly fine granularities. 
[Though our motivation radically differs from theirs, the following proposition shows that their approach belongs to the family of balanced weighted losses, and is thus a proper scoring rule.

\begin{proposition}\label{prop:bertinetto}
  The hierarchical cross-entropy loss \eqref{eq:bertinetto} can
  be rewritten in the form \eqref{eq:hierarchicalloss} with specific
  weights $w_j$ that obey the conditions of
  Proposition~\ref{prop:proper_loss}, and is thus a
  proper scoring rule.
\end{proposition}

\section{Experimental Protocol} \label{sec:4}

\subsection{Hierarchical Measures of Accuracy} \label{sec:hier_acc}

A class hierarchy calls into question the validity of the standard accuracy as a
performance criterion, since the classes are no longer considered equidistant
(see for example \cite{deng2010does,deselaers2011visual}). Deng \etal
\cite{deng2010does} use the distance between pairs of classes as the height of
their lowest common ancestor in a semantic graph like WordNet, in order to
quantify the difficulty of an image dataset. This height is a distance that is also relevant for
qualifying the severity of errors between ground-truth labels and their
prediction \cite{bertinetto2020making}.
We use the average of these individual heights
is a measure of the overall performance of a model (the lower the better).

In addition to this hierarchical distance, we introduce here two new
evaluation tools for hierarchical supervised classification.
The first one is the coarsening accuracy curve that displays the accuracy \vs the
coarsening of the classification problem.
More precisely, given a threshold $\tau$ such that
$0\leq\tau<\nicefrac{1}{2}$, we define a sub-tree $T_\tau$ by bottom-up
pruning of tree $T$ at nodes $\ell$ furthest from the root:
\begin{equation*}
  \sum_{j\in a\p{p\p{\ell}}}w_j<\tau\leq\sum_{j\in a\p{\ell}}w_j
  \enspace.
\end{equation*}
The subtree $T_{\nicefrac{1}{2}}$ is the tree $T$ itself, and for $\tau$
sufficiently small, $T_{\tau}$ reduces to the root with its
children as leaves, and $T_0$ reduces to the root for which the accuracy of any classifier is $1$.

The second measure targets the softmax outputs, before decisions
are made; it relies on optimal transport on graphs (see~\cite{COTFNT}
for a general introduction).  
It computes the cost of transporting, through the
tree, the distribution of softmax outputs to the ground-truth
distribution that is supported by the true label only.
Following~\cite{evans_phylogenetic_2012,le_treesliced_2019}, given two
distributions $\mu$, $\nu$ supported on $\mathcal V$ and the metric on
$\mathcal V$ defined by $d_T$, the Wasserstein distance between them
can be written as follows
\begin{equation}
  \label{eq:4}
  W\p{\mu, \nu}=\sum_{v_j\in\mathcal V}w_j\abs{\mu\p{\Gamma\p{v_j}}-\nu\p{\Gamma\p{v_j}}}
  \enspace,
\end{equation}
with $\Gamma\p{v_j}$ the subtree of $T$ rooted at $v_j$.
In our case, 
we show in Appendix \ref{sec:proofs} that the Wasserstein distance reduces to
\begin{equation}\label{eq:wasserstein_distance}
  W\p{f\p{\bx;\btheta}, y}=\sum_{k=1}^Kf_k\p{\bx;\btheta}\cdot d_{T}\p{y, k}
  \enspace,
\end{equation}
where $f_k\p{\bx;\btheta}$ is the distribution of softmax outputs and
$y$ is the ground-truth label.
This distance takes into account the full vector of estimated
probabilities.  In particular, even if the decision is correct, it
induces a cost for probability masses assigned to incorrect classes;
if the decision is not correct, the probability mass assigned to $y$
may reduce the cost compared to the decision error.

\subsection{Architectures, Baselines and Datasets}

The proposed method has been evaluated on two architectures that differ by an order of magnitude in their number of parameters: ResNet50 \cite{he2016deep} is a standard convolutional network for transfer learning and domain adaptation (23 M parameters), and MobileNetV3-Small \cite{howard2019searching} is a much smaller network (2.9 M parameters), which is aimed at low computing resources.

We compared our approach to two baselines: the standard fine-tuning strategy with cross-entropy, and with the hierarchical cross-entropy, which is the closest relative of our method in hierarchical supervised classification. 
These comparisons were carried out on three computer vision benchmarks.

\paragraph{iNaturalist19} \cite{van2018inaturalist}
was presented as a challenge for the FGVC6 workshop at CVPR 2019. 
It contains a total of 268,243 images with a maximum size of 800 pixels, representing 1,010 species, some being  highly similar, with highly diverse capture conditions.
The class hierarchy is the taxonomy of the species represented, which has a height of 8.
We used the same ratios as Bertinetto \etal \cite{bertinetto2020making} to split the dataset in training, validation and test sets, with 70\%, 15\% and 15\% of the images respectively.
Fine-tuning on this data set using networks pre-trained on ImageNet-1K is a transfer learning task.

\paragraph{TinyImageNet} \cite{chrabaszcz2017downsampled} is a subset of ImageNet with images downsampled to $64 \times 64$ pixels.
We used the version with 200 classes out of the 1,000 ones of ImageNet-1K, with 500 training samples per classes.
We built the class hierarchy from their class names using WordNet, leading to about the same depth than the hierarchy used for ImageNet-1K. 
Some class names have been edited by hand to avoid confusion with a homonym.
Fine-tuning on this data set using networks pre-trained on ImageNet-1K is a domain adaptation task.

\paragraph{ImageNet-1K} \cite{imagenet} contains 1,281,167 training images and 50,000 validation images representing 1,000 object classes.
The class hierarchy, taken from \cite{liu_large_2019}, has height 14.
We downsampled the original images to $64 \times 64$, making it a larger TinyImageNet with more classes and samples. 
Fine-tuning on these images using networks pre-trained on ImageNet-1K simulates a domain adaptation task.

We test our approach in several data regimes for each benchmark, varying the number of training images per class, from a few to the entire training data set.
All experiments were repeated three times, on different training subsets, in order to estimate the variability of the learning process. 
For iNaturalist, we used 1/32nd, 1/16th, 1/8th, 1/4th 1/2th and the entire 
training set;
for TinyImageNet, we used 1/32nd, 1/16th, 1/8th, 1/4th 1/2th and the entire 
training set;
for ImageNet, we used 1/128th, 1/64th, 1/32nd, 1/8th, 1/4th, 1/2th and the entire 
training set.
The training subsets used are the same for all the methods compared and were generated randomly.

\subsection{Training Protocol}

As is typical for small training sets, all networks are fine-tuned from pre-trained parameters.
We use the latest versions of the ImageNet pre-trained models supplied by Torchvision: V2 for ResNet50 and V1 for MobileNetV3-Small.
All the images of iNaturalist19, TinyImageNet and our downsampled version of ImageNet were then scaled up by bilinear interpolation to match the input sizes of the pre-trained architectures, that is, $224\times224$.

We fine-tune the networks using the Adam optimizer \cite{kingma2014adam} with a cosine annealing scheduler with a period equal to the total number of epoch. The initial learning rate is set to $10^{-4}$.
We use the AutoAugment \cite{Cubuk_2019_CVPR} implementation from PyTorch with the ImageNet policy for data augmentation. 
No weight decay was required to avoid overfitting.

HXE \cite{bertinetto2020making} requires the adjustment of a hyper-parameter which we have selected from the range of values considered in the original article. We chose $\alpha = 0.1$ (see Equation~\ref{eq:bertinetto}), which leads to the best performance in terms of accuracy , and $\alpha = 0.5$, which lead to the best compromise for lower the amount of gross mistakes in \cite{bertinetto2020making}.
Our method also needs to set the growth rate $q$ that defines the tree weighting used during learning. 
We report here the results with $q=1.2$ and $q=0.9$. Settings with $q>1$ give more weight to the leaves, thus avoiding over-focusing on superclasses and overlooking fine-grained classes while settings with $q<1$ give more weights to the superclasses.

\section{Results and Analysis} \label{sec:5}

\subsection{Variable Training Set Size}

\begin{figure} 
  \tiny
  \centering
  \begin{tabular}{@{}r@{\ \ }r@{\ }|@{\ }c@{\ \ \ }c@{\ \ \ }c@{}}
     & & \multicolumn{1}{c|}{\small{Wasserstein Dist.}} & \multicolumn{1}{c|}{\small{Hierarchical Dist.}} & \multicolumn{1}{c|}{\small{Accuracy}}\\
     \cline{1-5}
	 \multirow{2}{*}{\rotatebox[origin=l]{90}{\makebox[0cm][c]{\small{iNaturalist}}}} 
     & \rotatebox[origin=l]{90}{\makebox[2.5cm][c]{\small{ResNet}}} &
     \includegraphics[width=0.28\textwidth,,trim={0.7cm 0.7cm 5.5cm 0cm},clip]{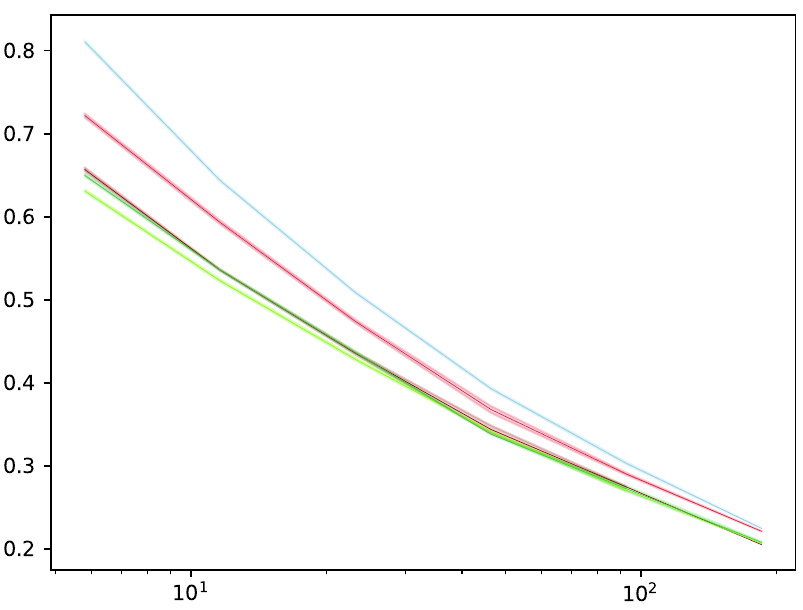} & 
	 \includegraphics[width=0.28\textwidth,,trim={0.7cm 0.7cm 5.1cm 0.1cm},clip]{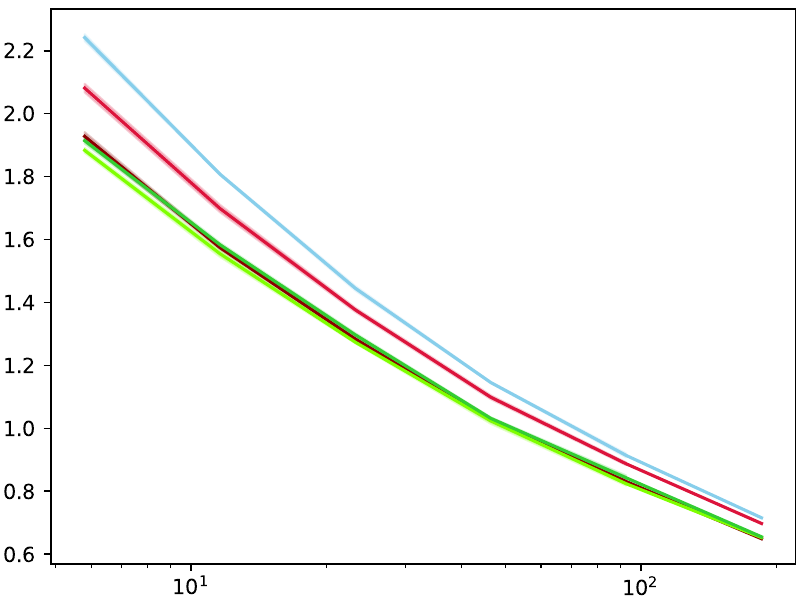} & 
     \includegraphics[width=0.28\textwidth,,trim={0.7cm 0.7cm 5.5cm 0cm},clip]{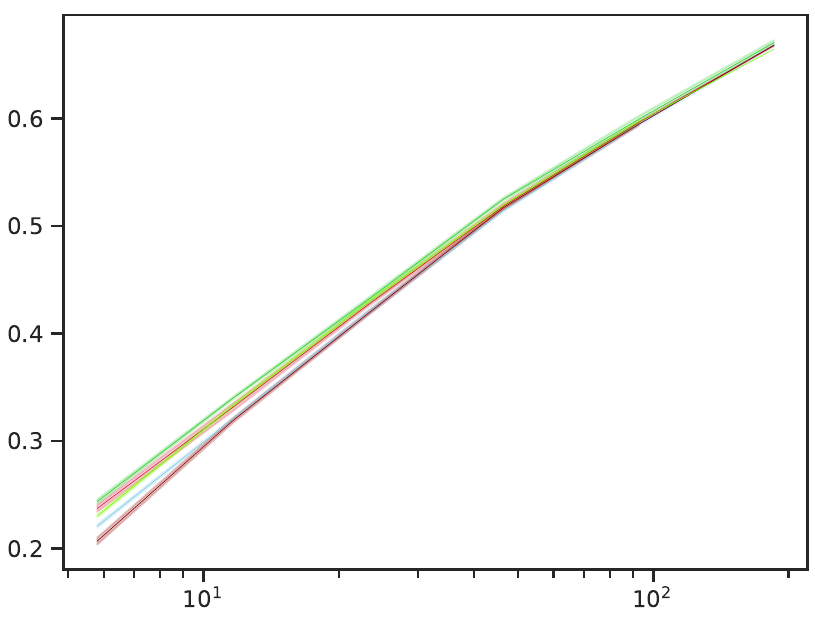} \\ 
	 \cline{2-2} \\[-0.2cm]
     & \rotatebox[origin=l]{90}{\makebox[2.5cm][c]{\small{MobileNet}}} &
     \includegraphics[width=0.28\textwidth,,trim={0.7cm 0.7cm 5.5cm 0cm},clip]{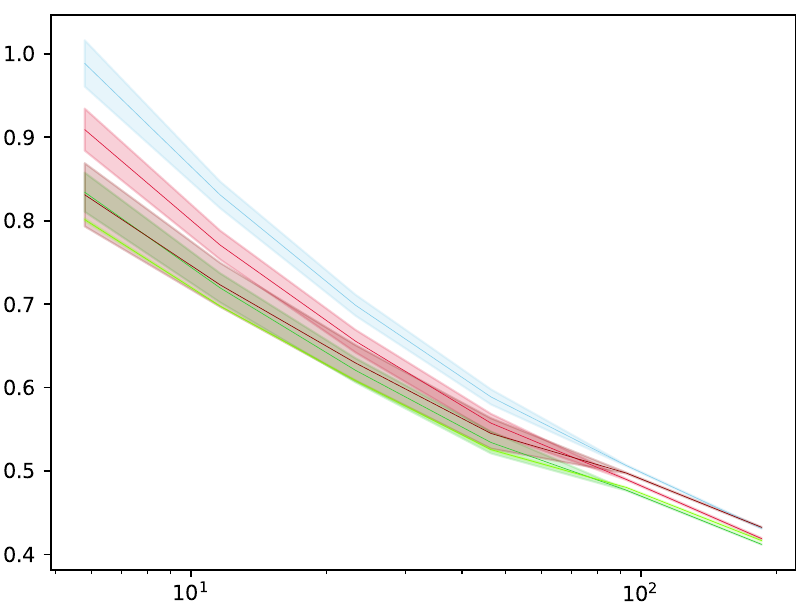} & 	
     \includegraphics[width=0.28\textwidth,,trim={0.7cm 0.7cm 5.5cm 0cm},clip]{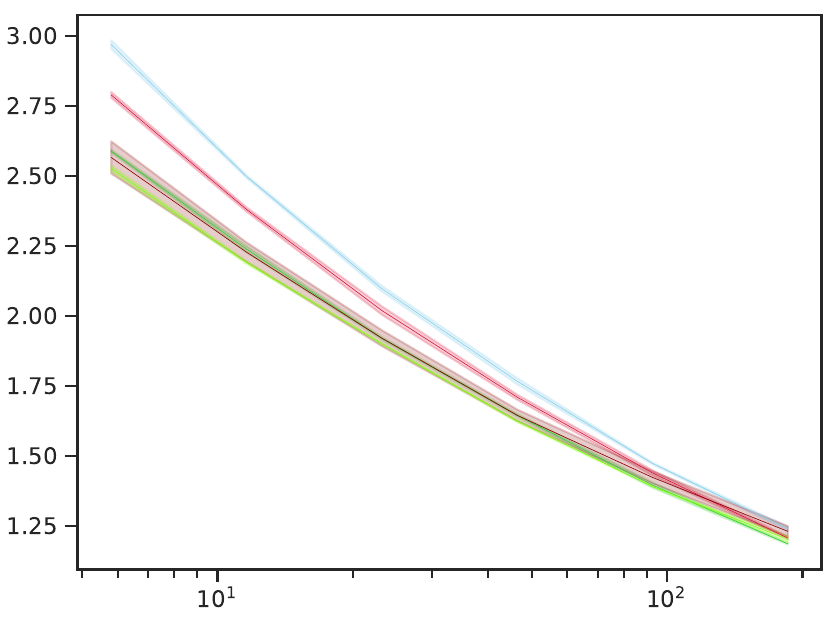} & 
     \includegraphics[width=0.28\textwidth,,trim={0.7cm 0.7cm 5.5cm 0cm},clip]{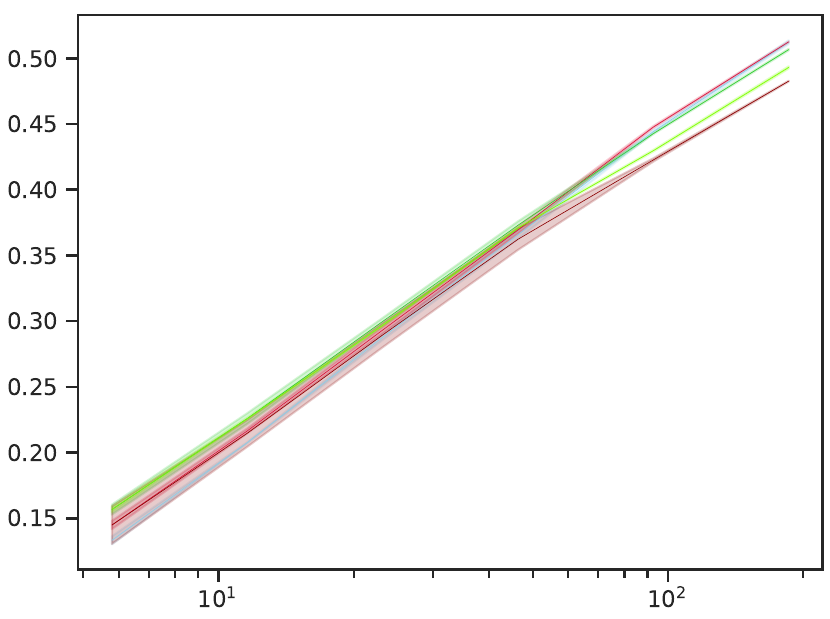}  \\
     \cline{1-2} \\[-0.2cm]
	 \multirow{2}{*}{\rotatebox[origin=l]{90}{\makebox[0cm][c]{\small{TinyImageNet}}}} 
     & \rotatebox[origin=l]{90}{\makebox[2.5cm][c]{\small{ResNet}}} &
     \includegraphics[width=0.28\textwidth,,trim={0.7cm 0.7cm 5.5cm 0cm},clip]{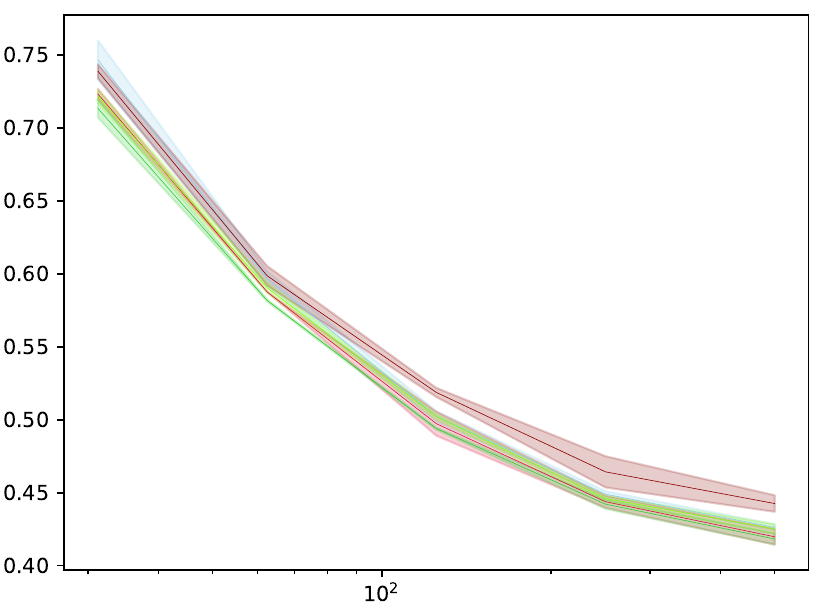} & 
	 \includegraphics[width=0.28\textwidth,,trim={0.7cm 0.7cm 5.1cm 0cm},clip]{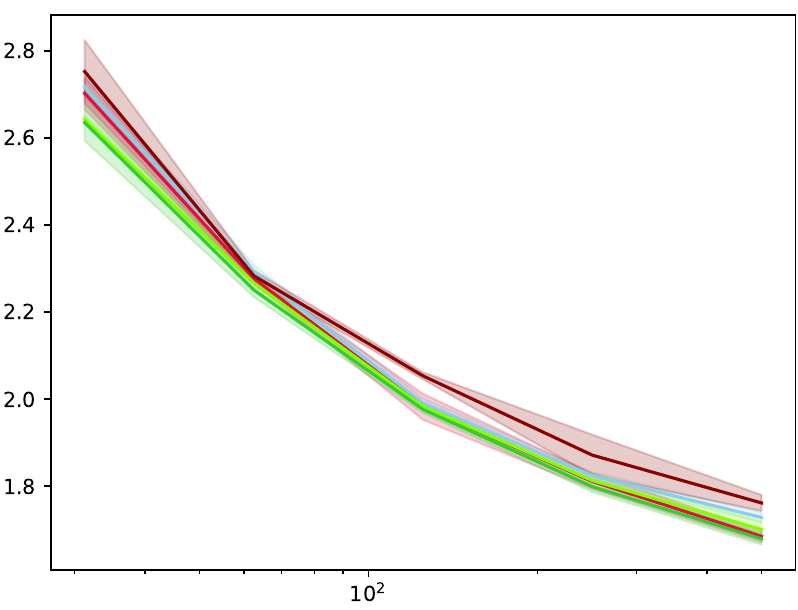} & 
     \includegraphics[width=0.28\textwidth,,trim={0.7cm 0.7cm 5.5cm 0cm},clip]{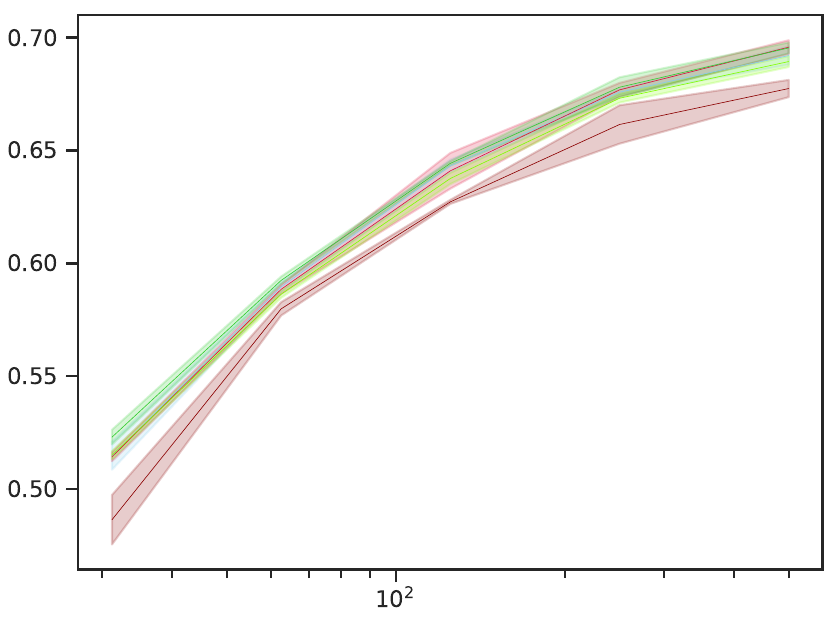} \\
	 \cline{2-2} \\[-0.2cm]
     & \rotatebox[origin=l]{90}{\makebox[2.5cm][c]{\small{MobileNet}}} &
     \includegraphics[width=0.28\textwidth,,trim={0.7cm 0.7cm 5.5cm 0cm},clip]{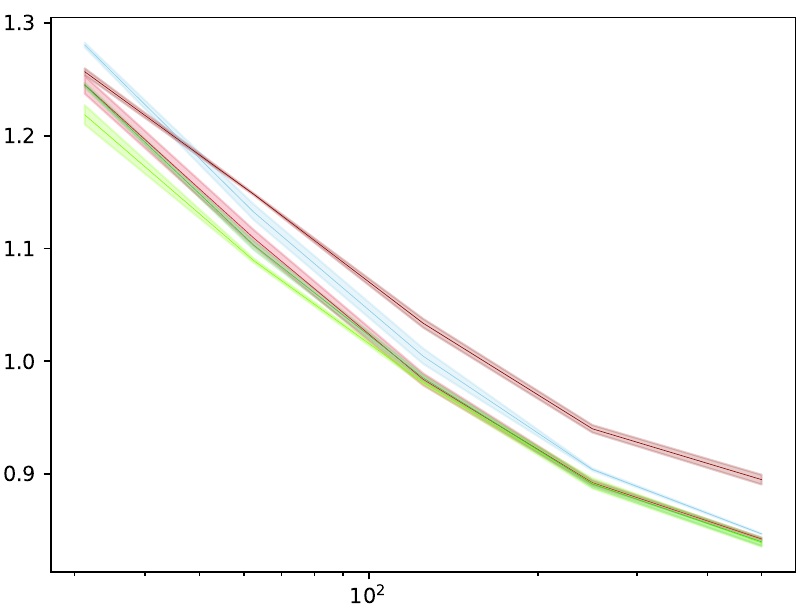} & 	
     \includegraphics[width=0.28\textwidth,,trim={0.7cm 0.7cm 5.1cm 0cm},clip]{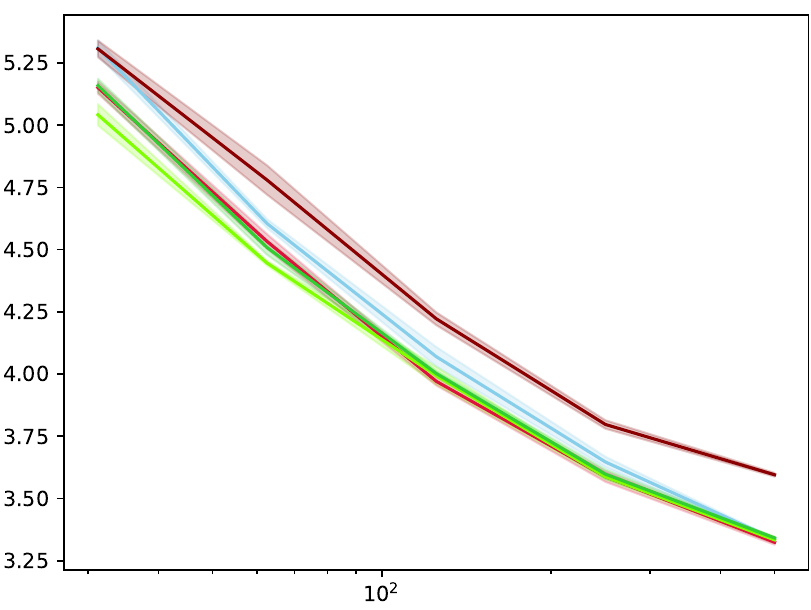} & 
     \includegraphics[width=0.28\textwidth,,trim={0.7cm 0.7cm 5.5cm 0cm},clip]{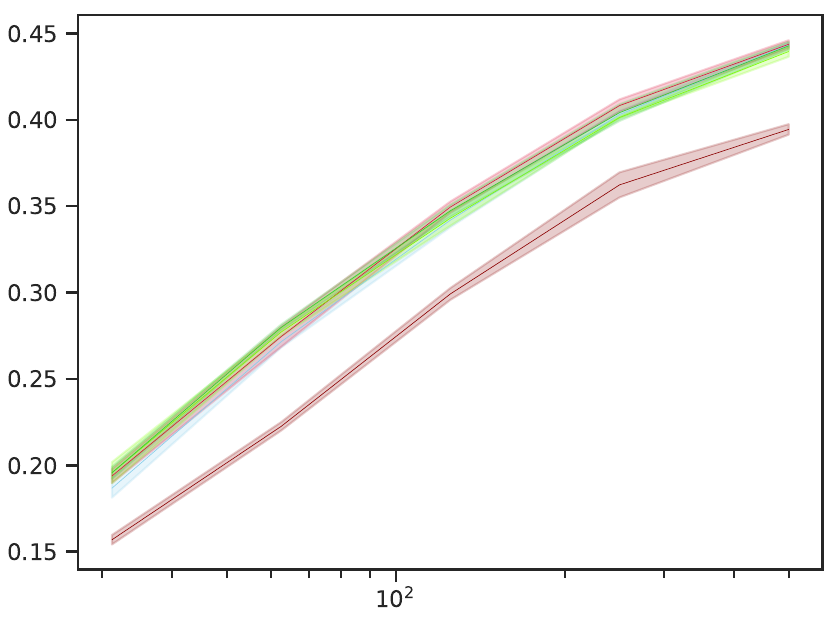} \\
     \cline{1-2} \\[-0.2cm]
	 \multirow{2}{*}{\rotatebox[origin=l]{90}{\makebox[0cm][c]{\small{ImageNet}}}} 
     & \rotatebox[origin=l]{90}{\makebox[2.5cm][c]{\small{ResNet}}} &
     \includegraphics[width=0.28\textwidth,,trim={0.7cm 0.7cm 5.5cm 0cm},clip]{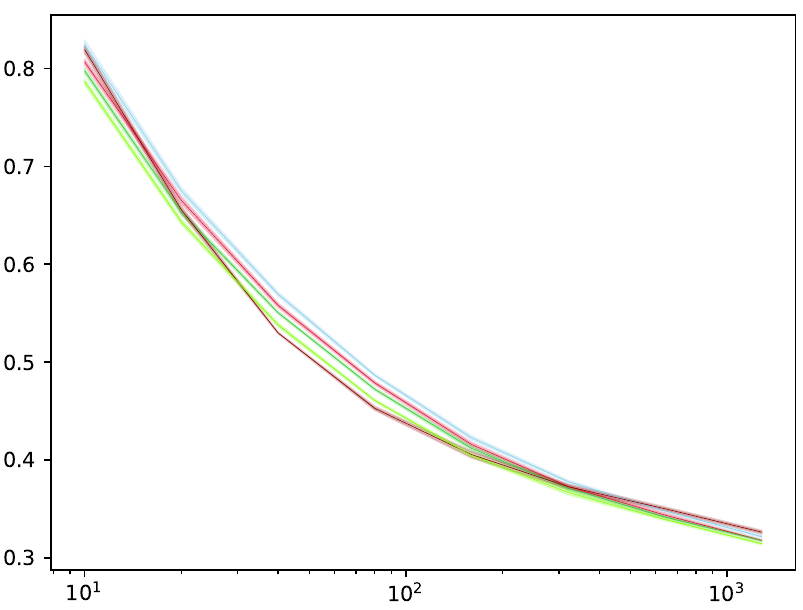} & 
	 \includegraphics[width=0.28\textwidth,,trim={0.7cm 0.7cm 5.1cm 0cm},clip]{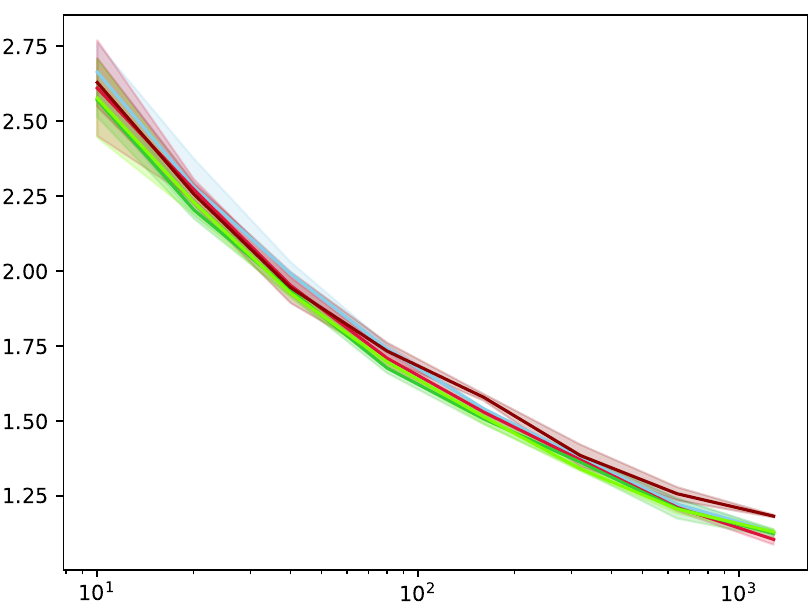} & 
     \includegraphics[width=0.28\textwidth,,trim={0.7cm 0.7cm 5.5cm 0cm},clip]{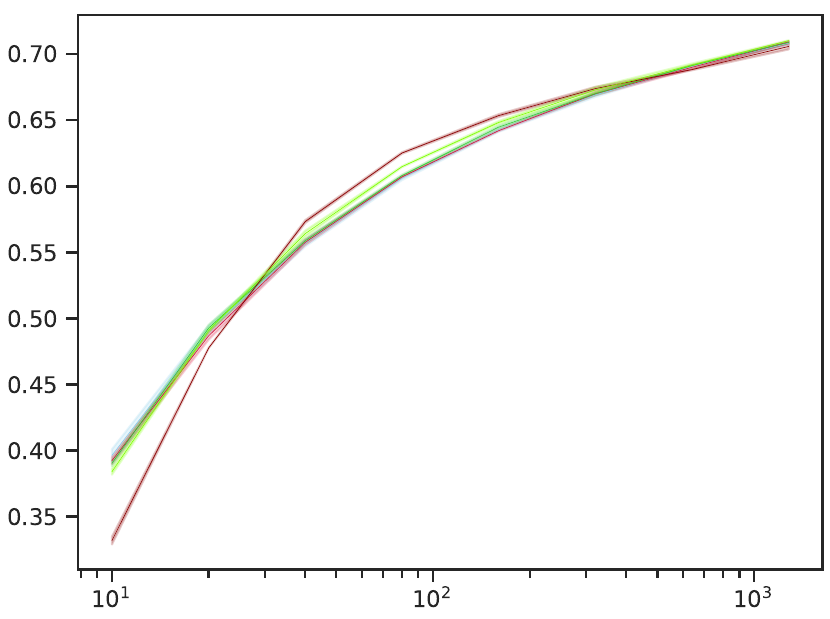} \\
	 \cline{2-2} \\[-0.2cm]
     & \rotatebox[origin=l]{90}{\makebox[2.5cm][c]{\small{MobileNet}}} &
     \includegraphics[width=0.28\textwidth,,trim={0.7cm 0.7cm 5.5cm 0cm},clip]{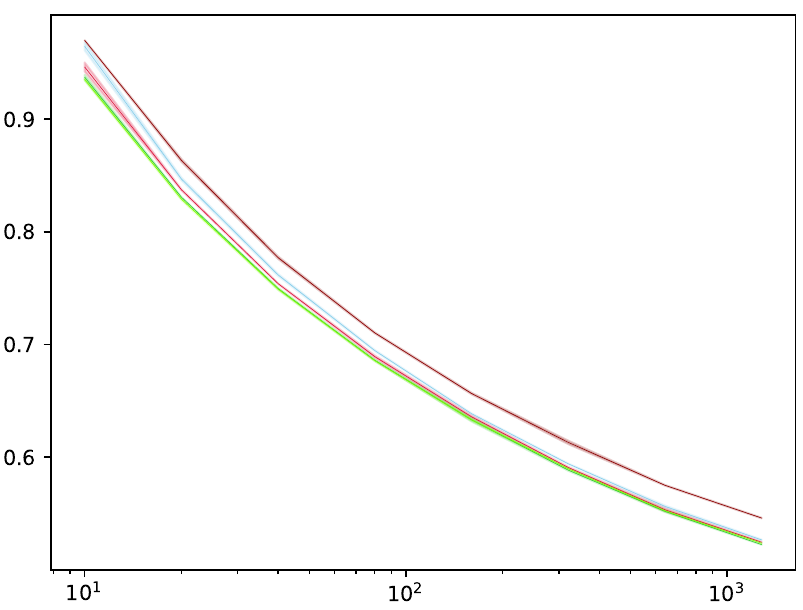} & 	
     \includegraphics[width=0.28\textwidth,,trim={0.7cm 0.7cm 5.1cm 0cm},clip]{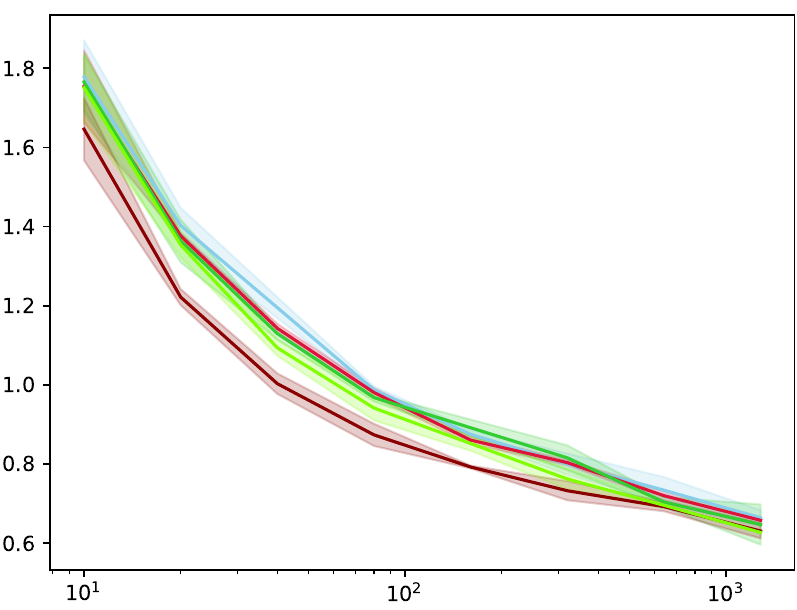} & 
     \includegraphics[width=0.28\textwidth,,trim={0.7cm 0.7cm 5.5cm 0cm},clip]{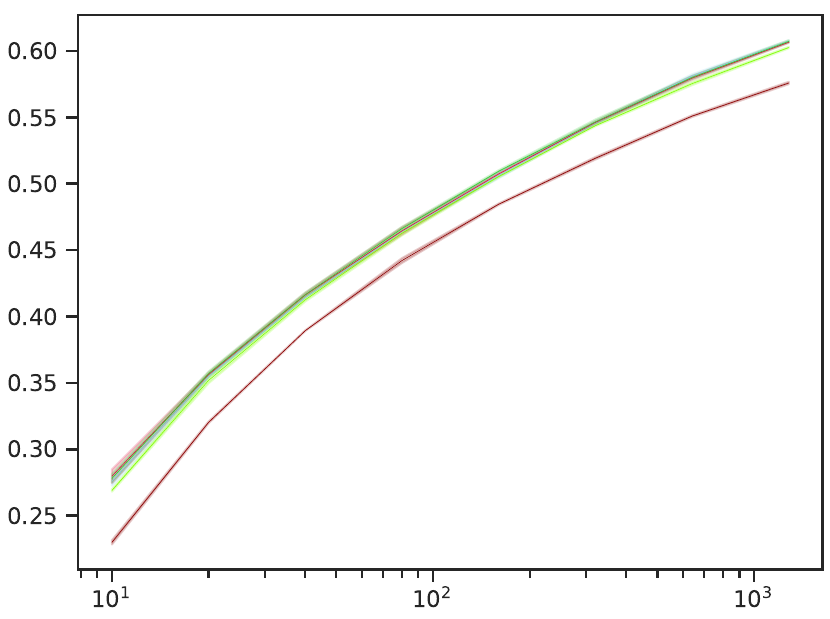}  \\[-1ex]
     \cline{1-2} \\[-0.2cm]
     & \multicolumn{2}{@{}c@{}} {\# training samples per class} & 
     \# training samples per class & 
     \# training samples per class \\[-1ex]
   \end{tabular}

   \medskip
   \small
   \begin{tabular}{*{5}{ll}}
	{\textcolor{crossentropy}{\rule[0.5ex]{1.5em}{1pt}}} &  { Cross-entropy} &
    {\textcolor{bertinetto1}{\rule[0.5ex]{1.5em}{1pt}}} & { HXE, $\alpha = 0.1$} &
    {\textcolor{bertinetto5}{\rule[0.5ex]{1.5em}{1pt}}} & { HXE, $\alpha = 0.5$} \\
	& & {\textcolor{hier9}{\rule[0.5ex]{1.5em}{1pt}}}  & { Ours, $q=0.9$} &
	{\textcolor{hier12}{\rule[0.5ex]{1.5em}{1pt}}}  & { Ours, $q=1.2$} 
  \end{tabular}

	  \caption{Average Wasserstein distance, hierarchical distance and standard accuracy for the two backbones as a function of the number of training images for the three benchmarks. 
   The lines represent the average results on the replicates, and the colored areas represent $\pm1$ standard deviation.}
   \label{fig:results_all}
\end{figure}

Figure~\ref{fig:results_all} summarizes the results when varying the training
set sizes for the three benchmarks and the two backbones. As expected,
MobileNetV3-Small is outperformed by ResNet50 regarding all performance measures
and for all training strategies, but the curves for MobileNetV3-Small demonstrate that the observations concerning the relative performance of methods applied to a large network designed for performance are similar for small networks designed for computational efficiency with limited
resources.

We have the same overall trends for the two backbones and the three benchmarks, except for some additional
variability for TinyImageNet that has a smaller number of classes. For small training set sizes, our method and hierarchical
cross-entropy (HXE) are consistently better than
cross-entropy in terms of hierarchical distance. We see that this improvement is
not paid by a loss in standard accuracy since we obtain equivalent standard
accuracies.
The reduction in hierarchical distance is also reflected on the average Wasserstein
distance, which means that there is less predicted probability mass that is put far from the ground truth for the networks trained
with our approach.
For large datasets, there is little difference between the approaches. This is in line with our expectations, since the networks can then learn hierarchy-related regularities from the data.

On average, setting $q=0.9$ yields better scores on hierarchical metrics compared to $q=1.2$, while maintaining comparable or superior fine-class accuracies.
HXE with $\alpha=0.5$ is less accurate than with $\alpha=0.1$, which in line with the results of  \cite{bertinetto2020making}, with a few exceptions on ImageNet with ResNet. For the other experimental settings, it sometimes even underperforms cross-entropy. 
HXE with $\alpha=0.1$ gives classification accuracies comparable to our method, but is slightly worse on hierarchical metrics.

\subsection{Coarsening Accuracy Curves}

\begin{figure}[t] 
  \centering
  \small
  \begin{tabular}{@{}r@{\ }|c@{\ \ \ }c@{\ \ \ }c@{}}
     \multicolumn{1}{c}{} & \multicolumn{1}{c}{\small{iNaturalist}} & \multicolumn{1}{c}{\small{TinyImageNet}} & \multicolumn{1}{c}{\small{ImageNet}}\\
     & \multicolumn{1}{c|}{\small{6 samples per class}} & \multicolumn{1}{c|}{\small{32 samples per class}} & \multicolumn{1}{c|}{\small{20 samples per class}}\\
     \hline
     \rotatebox[origin=l]{90}{\makebox[2.5cm][c]{\small{ResNet50}}} &
     \includegraphics[width=0.28\textwidth,trim={0.7cm 0.7cm 5.5cm 0cm},clip]{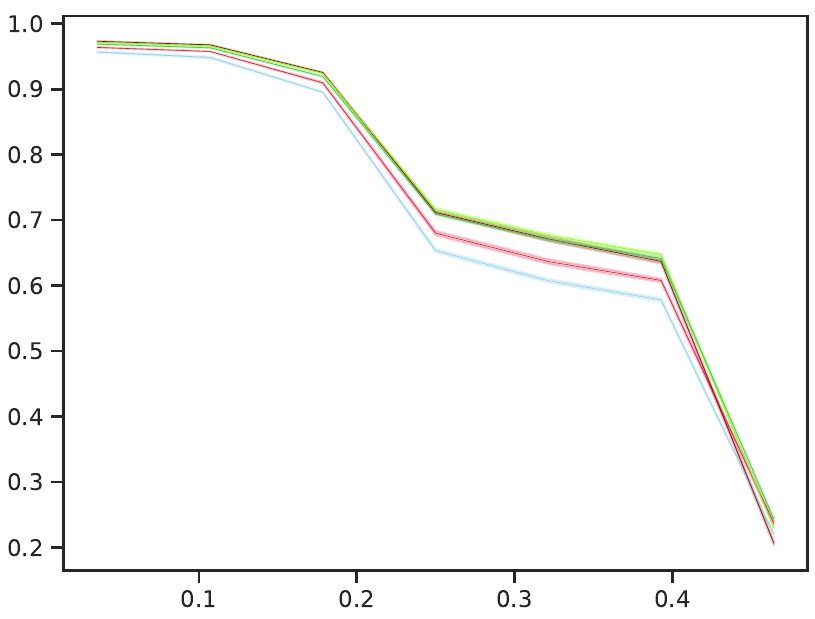} & 
	 \includegraphics[width=0.28\textwidth,trim={0.7cm 0.7cm 5.5cm 0cm},clip]{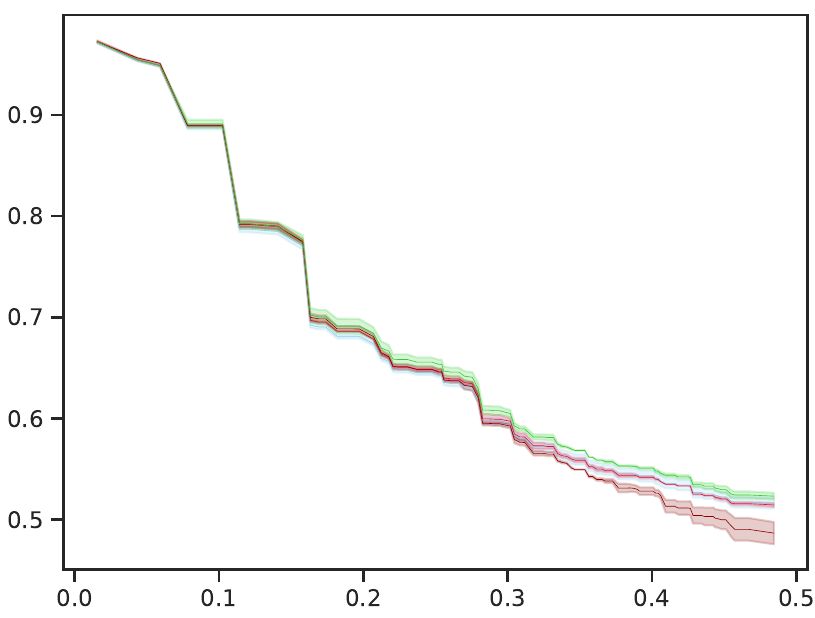} & 
     \includegraphics[width=0.28\textwidth,trim={0.7cm 0.7cm 5.5cm 0cm},clip]{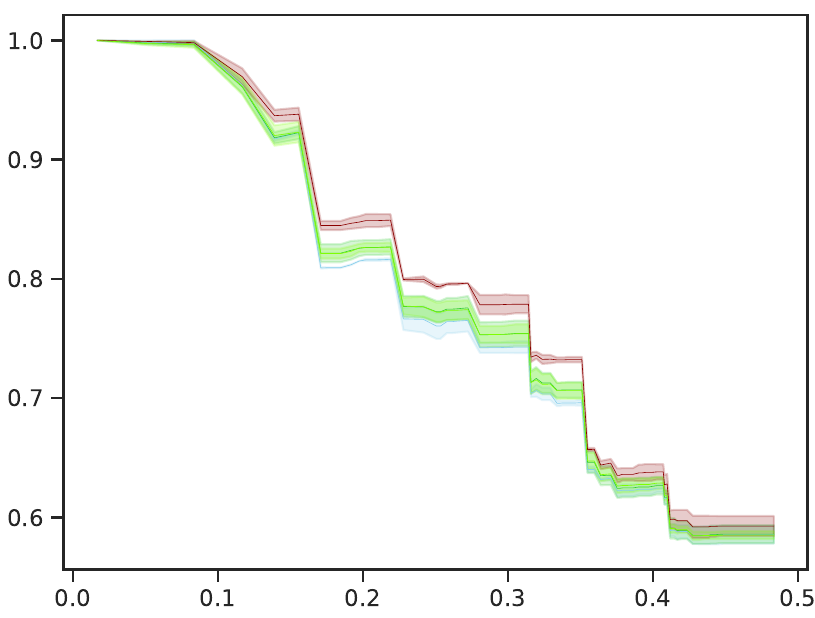} \\
	 \cline{1-1} \\[-0.2cm]
     \rotatebox[origin=l]{90}{\makebox[2.5cm][c]{\small{MobileNetV3}}} &
     \includegraphics[width=0.28\textwidth,trim={0.7cm 0.7cm 5.5cm 0cm},clip]{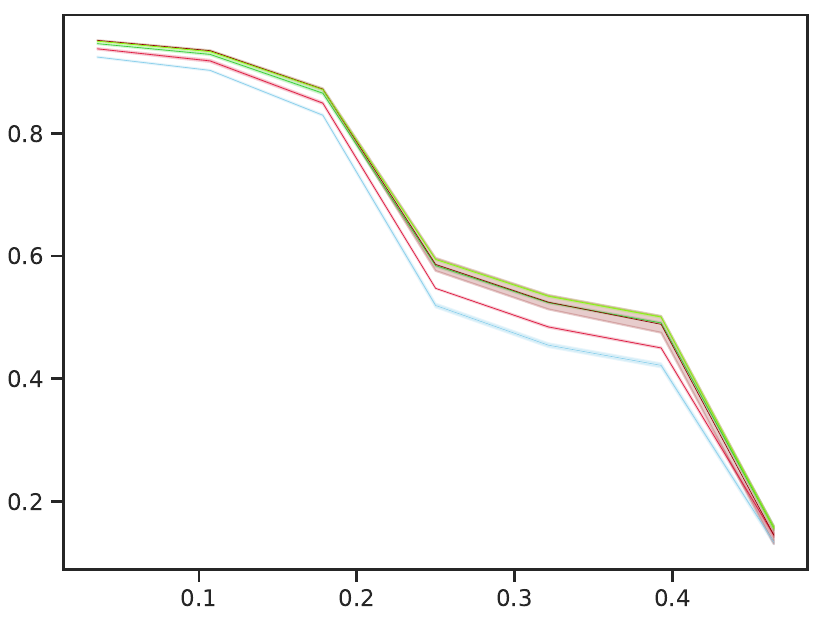} & 	
     \includegraphics[width=0.28\textwidth,trim={0.7cm 0.7cm 5.5cm 0cm},clip]{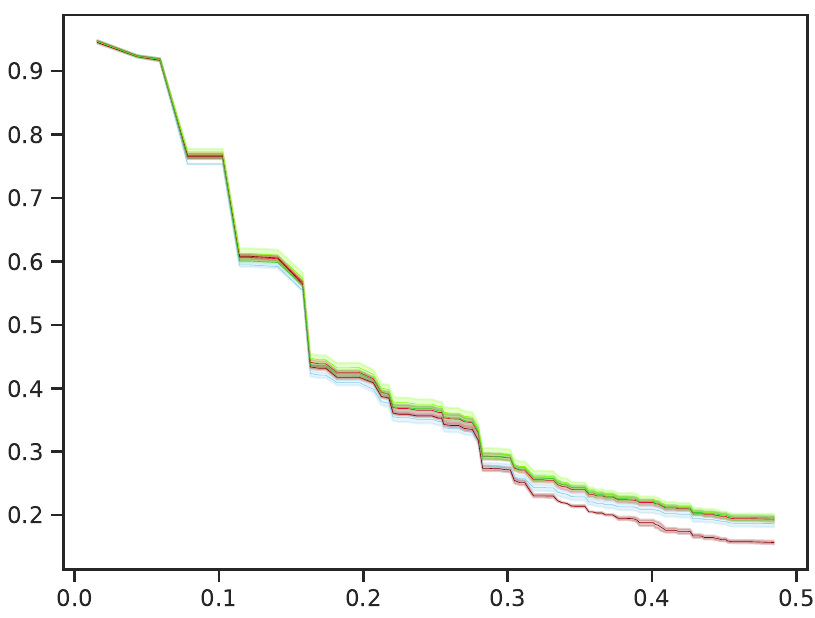} & 
     \includegraphics[width=0.28\textwidth,trim={0.7cm 0.7cm 5.5cm 0cm},clip]{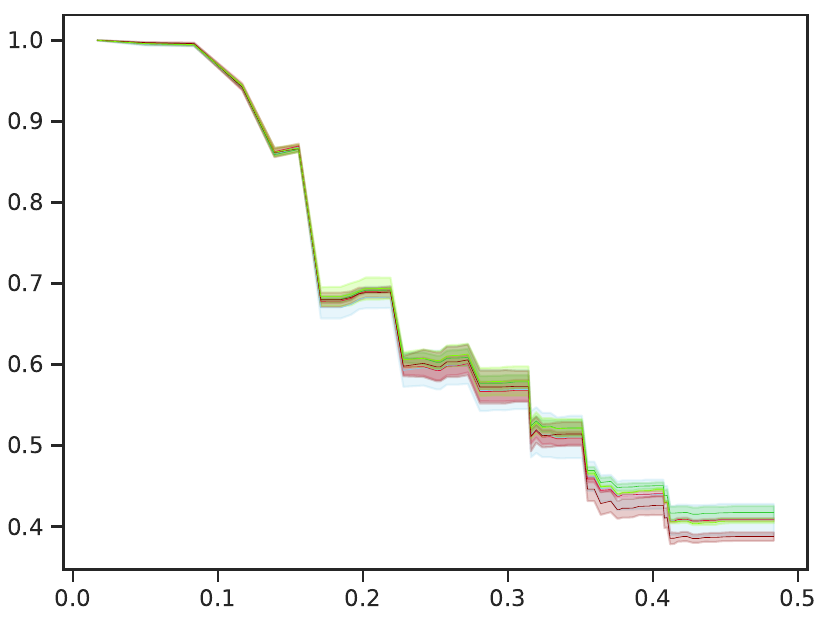}  \\
     \cline{1-1}
  \end{tabular} 
   
  \medskip
  \small
  \begin{tabular}{*{5}{ll}}
	 {\textcolor{crossentropy}{\rule[0.5ex]{1.5em}{1pt}}} &  { Cross-entropy} &
     {\textcolor{bertinetto1}{\rule[0.5ex]{1.5em}{1pt}}} & { HXE, $\alpha = 0.1$} &
     {\textcolor{bertinetto5}{\rule[0.5ex]{1.5em}{1pt}}} & { HXE, $\alpha = 0.5$} \\
	 & & {\textcolor{hier9}{\rule[0.5ex]{1.5em}{1pt}}}  & { Ours, $q=0.9$} &
	 {\textcolor{hier12}{\rule[0.5ex]{1.5em}{1pt}}}  & { Ours, $q=1.2$} 
  \end{tabular}
    \caption{Coarsening accuracy curves of accuracies \vs coarsening of the classification problem (the higher the better), for the two backbones and the three benchmarks. The lines represent the average results on the replicates, and the colored areas represent $\pm1$ standard deviation. }
    \label{fig:counts:threshold}
\end{figure}

Figure \ref{fig:counts:threshold} shows the Coarsening Accuracy Curves, as defined in Section \ref{sec:hier_acc}.
The iNaturalist class tree is balanced, with depth 8, so there are only 7 thresholds corresponding to different classifications.
The trees for TinyImageNet and ImageNet are deeper and unbalanced, so there are many more thresholds corresponding to different classifications.
These graphs are computed by aggregating the probabilities of the classes corresponding to each superclass,  and therefore show the benefit of not scattering the probability masses far from the ground-truth label.

On configurations where our method brings improvements on hierarchical metrics, the accuracies are higher, especially for moderate coarsening, with thresholds that are closer to $\nicefrac{1}{2}$. With thresholds close to zero, the classification problem becomes simple, with less classes that are semantically farther apart, and the classification accuracy obtained with cross-entropy approaches 1. The simplification of the class hierarchy leaves little room for improving classification through knowledge of superclasses.

For the iNaturalist and TinyImageNet datasets, with both backbones, our method dominates cross-entropy and HXE 
For ImageNet with the ResNet50 model, HXE dominates the other methods, and there is a slight advantage for ours with MobileNetV3.

\subsection{Sensitivity to Hyper-Parameters}

Our method includes a hyper-parameter $q$  that controls the relative weight of superclasses to fine-grained classes.
Even if all the classification objectives are consistent, this relative weight has an impact on the results, and the same observation applies to HXE with the hyper-parameter $\alpha$.

\begin{figure}[t]
    \centering    
    \scriptsize
    \begin{tabular}{lc@{\ \ }l}
     \rotatebox[origin=l]{90}{\makebox[0cm][c]{\small{Wasserstein dist.}}} &
    \raisebox{-.5\height}{\includegraphics[width=0.43\textwidth,,trim={0.7cm 0.7cm 5.0cm 0cm},clip]{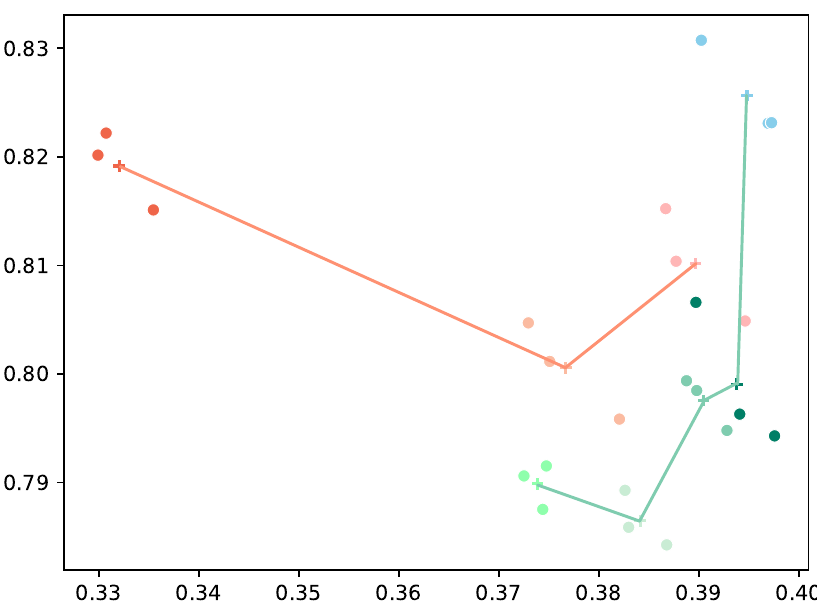}} &
   \begin{tabular}{ll}
     \footnotesize
     {\color[HTML]{87CCEB} $\bullet$} & Cross-entropy \\ 
     {\color[HTML]{FFB6B6} $\bullet$} & HXE, $\alpha = 0.1$ \\ 
     {\color[HTML]{FBBBA1} $\bullet$} & HXE, $\alpha = 0.3$ \\ 
     {\color[HTML]{FE9172} $\bullet$} & HXE, $\alpha = 0.4$ \\ 
     {\color[HTML]{EF6549} $\bullet$} & HXE, $\alpha = 0.5$ \\ 
     {\color[HTML]{DE3C24} $\bullet$} & HXE, $\alpha = 0.6$ \\ 
     {\color[HTML]{8FFFAC} $\bullet$} & Ours, $q=0.8$ \\ 
     {\color[HTML]{C9ECD4} $\bullet$} & Ours, $q=0.9$ \\ 
     {\color[HTML]{BEE0BF} $\bullet$} & Ours, $q=1.1$ \\ 
     {\color[HTML]{7FCCAF} $\bullet$} & Ours, $q=1.2$ \\ 
     {\color[HTML]{41B5A2} $\bullet$} & Ours, $q=1.3$ \\ 
     {\color[HTML]{007F66} $\bullet$} & Ours, $q=1.4$ \\ 
   \end{tabular} \\
  & Accuracy \\
  \end{tabular}

    \caption{%
      Wasserstein distance \vs classification accuracy with
      ResNet models trained on ImageNet with $6$ samples per class (best results in the bottom right-hand corner). Each point 
      represents a training run, and hues correspond to the hyper-parameter settings, for which the crosses represent the average results.
      The lines underline the evolution of the average results of the methods as a function of their hyperparameter, in red for HXE, in green for ours.
    }
    \label{fig:hyperparameters}
\end{figure}

Figure~\ref{fig:hyperparameters} illustrates the impact of these hyper-parameters ($q$ in our case and $\alpha$
for HXE).  It shows the results obtained on several runs with a ResNet model on a subset of ImageNet for a series of hyper-parameters in the plane formed by Wasserstein accuracy and validation accuracy.
Both methods converge to the standard cross-entropy as $q\rightarrow+\infty$ for our method and as $\alpha\rightarrow 0$ for HXE. 
We see that our method dominates HXE in the sense that for any parameter $\alpha$, there is a model with our weighting scheme that is strictly better on both metrics, towards the bottom right-hand corner.

\section{Conclusion} \label{sec:6}

Many techniques to exploit the knowledge of a hierarchy on classes have already been introduced \cite{5_xie_2015,4_koo_2020,10_yan_2015,9_goo_2016,15_wu_2020}, but few are suitable for use on any standard multiclass classifier.
Here we proposed a very simple generic approach for integrating the knowledge of a hierarchy over classes in learning, with a training loss that is automatically built from the class hierarchy. The approach is applicable to any model that is usually trained with cross-entropy. 

We show that the proposed loss is a proper scoring rule, meaning that its expectation is minimized by the true posterior probability. This property enables us to pursue improvements in coarse errors, without having to pay a price in terms of standard accuracy.

Our formulation defines a large and flexible family of weighting schemes for defining losses focusing more on the fine-grained or coarse-grained classification. 
Compared to standard cross-entropy, the computational cost of optimizing these losses is negligible.
Here, we chose to focus on weighting schemes that follow an exponential growth/decay along the tree path that goes from the set of classes to each individual class.
We show that the hierarchical cross-entropy of Bertinetto \etal \cite{bertinetto2020making}, although motivated by a radically different argument, based on the decomposition of posterior probabilities into conditional probabilities along the tree path, belongs to this family.

We conducted experiments with two networks on three benchmarks, with varying training sample sizes, that show that this type of losses are beneficial for small sample sizes, with a benefit in terms of hierarchical distance, and no degradation of the accuracy. The benefits are moderate, but they come at no price, and they are obtained on highly efficient models, with a strategy of data augmentation and a scheduler that make the baseline highly competitive. This strategy should be adopted when data is scarce.

As by-product summary statistics, we introduced a Wasserstein distance for evaluating the softmax outputs of a network model, by considering the transport of probability masses at the softmax layer through the tree of classes. We also introduce the coarsening accuracy curve that surveys in a single picture the accuracies that can be reached by considering the possible classification problems that are visited when coarsening the class hierarchy, by the bottom-up aggregation of classes.

In the future, we would like to use this approach for training with partially labelled examples that may either represent a partial knowledge of the data-labeling expert or an intrinsic uncertainty in the data. For example, if we have an image of an unknown dog breed, we can still use the knowledge of the superclass ``dog'' for training.
On the same line, but this time regarding predictions, we could deliver cautious partial predictions rather than bold precise ones. This requires calibrated predicted probabilities, and thanks to the proper scoring property, we know that our criterion should not hurt calibration. However, as deep learning is known to provide  poorly calibrated models~\cite{guo2017calibration}, delivering cautious coarsened predictions with these models is difficult.
 
\begin{credits}
\subsubsection{\ackname}
This work was carried out with the support of the French Agence Nationale de la Recherche through the ANR-20-CE10-0004 project.  
Access to IDRIS HPC resources has been granted under allocation AD011014519 made by GENCI.
\end{credits}

\bibliographystyle{splncs04}

\begin{thebibliography}{}
\providecommand{\url}[1]{\texttt{#1}}
\providecommand{\urlprefix}{URL }
\providecommand{\doi}[1]{https://doi.org/#1}

\end{thebibliography}


\begin{thebibliography}{10}
  \providecommand{\url}[1]{\texttt{#1}}
  \providecommand{\urlprefix}{URL }
  \providecommand{\doi}[1]{https://doi.org/#1}
  
  \bibitem{1_babbar_2016}
  Babbar, R., Partalas, I., Gaussier, E., Amini, M.R., Amblard, C.: Learning taxonomy adaptation in large-scale classification. J. Mach. Learn. Res.  \textbf{17}(98),  1--37 (2016)
  
  \bibitem{bertinetto2020making}
  Bertinetto, L., Mueller, R., Tertikas, K., Samangooei, S., Lord, N.A.: Making better mistakes: Leveraging class hierarchies with deep networks. In: Conf. Comput. Vis. Pattern Recog. ({CVPR}). pp. 12506--12515 (2020)
  
  \bibitem{blockeel_2002}
  Blockeel, H., Bruynooghe, M., D{\v{z}}eroski, S., Ramon, J., Struyf, J.: Hierarchical multi-classification. In: Workshop Notes of the KDD'02 Workshop on Multi-Relational Data Mining. pp. 21--35 (2002)
  
  \bibitem{Chang_2021_CVPR}
  Chang, D., Pang, K., Zheng, Y., Ma, Z., Song, Y.Z., Guo, J.: Your "flamingo" is my "bird": Fine-grained, or not. In: Conf. Comput. Vis. Pattern Recog. ({CVPR}). pp. 11476--11485 (June 2021)
  
  \bibitem{chrabaszcz2017downsampled}
  Chrabaszcz, P., Loshchilov, I., Hutter, F.: A downsampled variant of imagenet as an alternative to the {CIFAR} datasets. arXiv preprint  \textbf{abs/1707.08819} (2017)
  
  \bibitem{Cubuk_2019_CVPR}
  Cubuk, E.D., Zoph, B., Mane, D., Vasudevan, V., Le, Q.V.: {AutoAugment}: Learning augmentation strategies from data. In: Conf. Comput. Vis. Pattern Recog. ({CVPR}) (2019)
  
  \bibitem{deng2010does}
  Deng, J., Berg, A.C., Li, K., Fei-Fei, L.: What does classifying more than 10,000 image categories tell us? In: Eur. Conf. Comput. Vis. ({ECCV}). pp. 71--84. Springer (2010)
  
  \bibitem{7_deng_2014}
  Deng, J., Ding, N., Jia, Y., Frome, A., Murphy, K., Bengio, S., Li, Y., Neven, H., Adam, H.: Large-scale object classification using label relation graphs. In: Eur. Conf. Comput. Vis. ({ECCV}). pp. 48--64 (2014)
  
  \bibitem{imagenet}
  Deng, J., Dong, W., Socher, R., Li, L.J., Li, K., Fei-Fei, L.: Imagenet: A large-scale hierarchical image database. In: Conf. Comput. Vis. Pattern Recog. ({CVPR}). pp. 248--255 (2009)
  
  \bibitem{deselaers2011visual}
  Deselaers, T., Ferrari, V.: Visual and semantic similarity in {ImageNet}. In: Conf. Comput. Vis. Pattern Recog. ({CVPR}). pp. 1777--1784 (2011)
  
  \bibitem{evans_phylogenetic_2012}
  Evans, S.N., Matsen, F.A.: The phylogenetic {{Kantorovich}}–{{Rubinstein}} metric for environmental sequence samples. Journal of the Royal Statistical Society Series B: Statistical Methodology  \textbf{74}(3),  569--592 (2012)
  
  \bibitem{garg_eccv_2022}
  Garg, A., Sani, D., Anand, S.: Learning hierarchy aware features for reducing mistake severity. In: Avidan, S., Brostow, G., Ciss{\'e}, M., Farinella, G.M., Hassner, T. (eds.) Eur. Conf. Comput. Vis. ({ECCV}). pp. 252--267. Springer (2022)
  
  \bibitem{gneiting2007strictly}
  Gneiting, T., Raftery, A.E.: Strictly proper scoring rules, prediction, and estimation. Journal of the American statistical Association  \textbf{102}(477),  359--378 (2007)
  
  \bibitem{9_goo_2016}
  Goo, W., Kim, J., Kim, G., Hwang, S.J.: Taxonomy-regularized semantic deep convolutional neural networks. In: Eur. Conf. Comput. Vis. ({ECCV}). pp. 86--101 (2016). \doi{10.1007/978-3-319-46475-6\_6}
  
  \bibitem{guo2017calibration}
  Guo, C., Pleiss, G., Sun, Y., Weinberger, K.Q.: On calibration of modern neural networks. In: Int. Conf. Mach. Learn. ({ICML}). pp. 1321--1330 (2017)
  
  \bibitem{he2016deep}
  He, K., Zhang, X., Ren, S., Sun, J.: Deep residual learning for image recognition. In: Conf. Comput. Vis. Pattern Recog. ({CVPR}). pp. 770--778 (2016)
  
  \bibitem{howard2019searching}
  Howard, A., Sandler, M., Chu, G., Chen, L.C., Chen, B., Tan, M., Wang, W., Zhu, Y., Pang, R., Vasudevan, V., et~al.: Searching for {MobileNetV3}. In: Int. Conf. Comput. Vis. ({ICCV}). pp. 1314--1324 (2019)
  
  \bibitem{karthik2021no}
  Karthik, S., Prabhu, A., Dokania, P.K., Gandhi, V.: No cost likelihood manipulation at test time for making better mistakes in deep networks. In: Int. Conf. Learn. Represent. ({ICLR}) (2021)
  
  \bibitem{kingma2014adam}
  Kingma, D.P., Ba, J.L.: Adam: Amethod for stochastic optimization. In: Int. Conf. Learn. Represent. ({ICLR}) (2014)
  
  \bibitem{4_koo_2020}
  Koo, J., Klabjan, D., Utke, J.: Combined convolutional and recurrent neural networks for hierarchical classification of images. In: Int. Conf. on Big Data ({BigData}). pp. 1354--1361 (2020)
  
  \bibitem{le_treesliced_2019}
  Le, T., Yamada, M., Fukumizu, K., Cuturi, M.: Tree-sliced variants of {{Wasserstein}} distances. Adv. Neural Inform. Process. Syst. ({NeurIPS})  \textbf{32} (2019)
  
  \bibitem{liu_large_2019}
  Liu, Z., Miao, Z., Zhan, X., Wang, J., Gong, B., Yu, S.X.: Large-scale long-tailed recognition in an open world. In: Conf. Comput. Vis. Pattern Recog. ({CVPR}). pp. 2537--2546 (2019). \doi{10.1109/CVPR.2019.00264}
  
  \bibitem{nister2006scalable}
  Nister, D., Stewenius, H.: Scalable recognition with a vocabulary tree. In: Conf. Comput. Vis. Pattern Recog. ({CVPR}). vol.~2, pp. 2161--2168 (2006)
  
  \bibitem{COTFNT}
  Peyr\'e, G., Cuturi, M.: Computational optimal transport. Foundations and Trends in Machine Learning  \textbf{11}(5-6),  355--607 (2019)
  
  \bibitem{13_salakhutdinov_2011}
  Salakhutdinov, R., Torralba, A., Tenenbaum, J.B.: Learning to share visual appearance for multiclass object detection. In: Conf. Comput. Vis. Pattern Recog. ({CVPR}). pp. 1481--1488 (2011). \doi{10.1109/CVPR.2011.5995720}
  
  \bibitem{12_srivastava_2013}
  Srivastava, N., Salakhutdinov, R.: Discriminative transfer learning with tree-based priors. In: Adv. Neural Inform. Process. Syst. ({NeurIPS}). pp. 2094--2102 (2013)
  
  \bibitem{tanzi_hierarchical}
  Tanzi, L., Vezzetti, E., Moreno, R., Aprato, A., Audisio, A., Massè, A.: Hierarchical fracture classification of proximal femur x-ray images using a multistage deep learning approach. European Journal of Radiology  \textbf{133},  109373 (2020). \doi{https://doi.org/10.1016/j.ejrad.2020.109373}
  
  \bibitem{van2018inaturalist}
  Van~Horn, G., Mac~Aodha, O., Song, Y., Cui, Y., Sun, C., Shepard, A., Adam, H., Perona, P., Belongie, S.: The {iNaturalist} species classification and detection dataset. In: Conf. Comput. Vis. Pattern Recog. ({CVPR}). pp. 8769--8778 (2018)
  
  \bibitem{6_wang_2021}
  Wang, R., cai, D., Xiao, K., Jia, X., Han, X., Meng, D.: Label hierarchy transition: Modeling class hierarchies to enhance deep classifiers. arXiv preprint  \textbf{abs/2112.02353} (2021), \url{https://arxiv.org/abs/2112.02353}
  
  \bibitem{wu_hierarchical_2019}
  Wu, C., Tygert, M., LeCun, Y.: A hierarchical loss and its problems when classifying non-hierarchically. PLOS ONE  \textbf{14}(12),  1--17 (12 2019). \doi{10.1371/journal.pone.0226222}, \url{https://doi.org/10.1371/journal.pone.0226222}
  
  \bibitem{15_wu_2020}
  Wu, T., Morgado, P., Wang, P., Ho, C., Vasconcelos, N.: Solving long-tailed recognition with deep realistic taxonomic classifier. In: Eur. Conf. Comput. Vis. ({ECCV}). pp. 171--189 (2020). \doi{10.1007/978-3-030-58598-3\_11}
  
  \bibitem{11_wu_2017}
  Wu, Z., Saito, S.: {HiNet}: Hierarchical classification with neural network. arXiv preprint  \textbf{abs/1705.11105} (2017)
  
  \bibitem{5_xie_2015}
  Xie, S., Yang, T., Wang, X., Lin, Y.: Hyper-class augmented and regularized deep learning for fine-grained image classification. In: Conf. Comput. Vis. Pattern Recog. ({CVPR}). pp. 2645--2654 (2015). \doi{10.1109/CVPR.2015.7298880}
  
  \bibitem{10_yan_2015}
  Yan, Z., Zhang, H., Piramuthu, R., Jagadeesh, V., DeCoste, D., Di, W., Yu, Y.: {HD-CNN:} hierarchical deep convolutional neural networks for large scale visual recognition. In: Int. Conf. Comput. Vis. ({ICCV}). pp. 2740--2748 (2015)
  
  \bibitem{8_zhao_2011}
  Zhao, B., Fei{-}Fei, L., Xing, E.P.: Large-scale category structure aware image categorization. In: Adv. Neural Inform. Process. Syst. ({NeurIPS}). pp. 1251--1259 (2011)
  
  \bibitem{3_zhu_2017}
  Zhu, X., Bain, M.: {B-CNN:} branch convolutional neural network for hierarchical classification. arXiv preprint  \textbf{abs/1709.09890} (2017)
  
  \end{thebibliography}

\appendix

\section{Proofs}\label{sec:proofs}

\renewcommand*{\proofname}{Proof of Proposition \ref{prop:1}}
\begin{proof}
  For the first part of the proposition, we state that, for any leaf $v_{k}$, we
  have:
  \begin{equation*}
    \sum_{j\in a\p{k}} w_j= w_k+ \frac{1}{2} - \p{\frac{1}{2} - \sum_{j\in a\p{p\p{k}}} w_j}= w_k + \frac{1}{2} - w_k\frac{1-q^{h\p{{k}}+1}}{1-q} = \frac{1}{2} \enspace,
  \end{equation*}
  where we used Equation~\eqref{eq:2} and $h\p{k}=0$. 
  Thus weighting scheme produces a balanced weighted tree, where
  the weights along the path from the root to the leaves sum up to
  \nicefrac{1}{2}.

  For the second part of the proposition, we state that, for any non-root node
  $v_{j}$, we have:
  \begin{align*}
    w_j& = \p{\frac{1}{2} - \sum_{j\in a\p{p\p{j}}} w_j} \frac{1-q}{1-q^{h\p{j}+1}} \\
       &=  \p{ - w_{p\p{j}} + \frac{1}{2} - \sum_{j\in a\p{p\p{p\p{j}}}} w_j}\frac{1-q}{1-q^{h\p{j}+1}} \\
       & =  q \, \frac{1-q^{h\p{p\p{j}}}}{1-q^{h\p{j}+1}} w_{p\p{j}}
       \enspace,
    \end{align*}
    where Equation~\eqref{eq:2} is used twice. We remark that as soon as
    $h(p(j))=h(j)+1$, we have $w_{j}=qw_{p(j)}$.
\end{proof}

\renewcommand*{\proofname}{Proof of Proposition \ref{prop:proper_loss}}
\begin{proof}
  Let $\bx$ be an arbitrary value, and $\pi_{k}=\Proba(Y=k\mid
  X=\bx)$.  One has to prove that $\boldsymbol{\pi}=(\pi_{1},\ldots\pi_{K})$ minimizes
  the expected loss $\E_{Y \mid X}\p{\loss_{T}\p{\boldsymbol{\rmf},Y}}$ with respect
  to $\boldsymbol{\rmf}= (\rmf_{1},\ldots,\rmf_{K})$.  
  We have
  \begin{align*}
    \E_{Y \mid X}\p{\loss_{T}\p{\boldsymbol{\rmf},Y}}
	  &= \sum_{k=1}^K \pi_k \, \loss_{T}\p{\boldsymbol{\rmf}, k} \\
    & =-\sum_{k=1}^K \pi_k \sum_{j\in{a\p{k}}} w_j \log\Bigl( \sum_{\ell\in v_{j}} \rmf_{\ell}\Bigr) \\
      &=-\sum_{j} w_j \Bigl(\sum_{k\in v_{j}} \pi_k \Bigr) \log\Bigl( \sum_{\ell\in v_{j}} \rmf_{\ell}\Bigr)
	  \enspace,
  \end{align*}
  where the index $j$ covers all the nodes of the tree $T$ except the root, and where we use that $k$ is an element of all the ancestors of $k$. 
  It suffices to show that the gradient of the expected loss with respect to $\boldsymbol{\rmf}$
  along the probability simplex at point $\boldsymbol{\pi}$ vanishes, which is equivalent to
  show it is proportional to $\ones$, where $\ones$ is a vector of ones. 
  Indeed, we have
  \begin{align*}
    \frac{\partial\E_{Y \mid X}\p{\loss_T\p{\boldsymbol{\rmf}, Y}}}{\partial \rmf_{k}}
	&=\sum_{j\in a\p{k}} w_j \Bigl(\sum_{\ell\in v_{j}} \pi_\ell \Bigr) \frac{1}{\sum_{\ell\in v_{j}} \rmf_{\ell}}
	\enspace.
  \end{align*}
  When evaluating at point $\boldsymbol{\pi}$, $\displaystyle{\frac{\partial\E_{Y \mid X}\p{\loss_{T}\p{\boldsymbol{\rmf}, Y}}}{\partial \rmf_{k}}\Bigr|_{\boldsymbol{\rmf}=\boldsymbol{\pi}} = \sum_{j\in a\p{k}}w_j}$, \\
  which is constant for all $k\in\{1,\ldots,K\}$ if and only if the weighting is balanced.
\end{proof}

\renewcommand*{\proofname}{Proof of Proposition \ref{prop:not_proper_loss}}
\begin{proof}
  Proposition \ref{prop:not_proper_loss} is a corollary of Proposition~\ref{prop:proper_loss} with $w_{j}=1$.
\end{proof}

\renewcommand*{\proofname}{Proof of Proposition \ref{prop:bertinetto}}
\begin{proof} 
	The hierarchical cross-entropy loss $\loss_{\text{HXE}}$ \eqref{eq:bertinetto} reads:
	\begin{equation}
	     \lambda\p{d\p{y}}
		 \log\p{f_{y}(\bx;\btheta)} - %
		 \sum_{j\in{a\p{p(y)}}} 
		 \p{\lambda\p{d\p{j}}-\lambda\p{d\p{j}+1}} 
		 \log\p{\sum_{k\in v_{j}} f_{k}(\bx;\btheta)} 
	     \enspace, \nonumber
     \end{equation}
	 where $\lambda\p{d\p{j}}=\exp\p{-\alpha d\p{j}}$ and $d\p{j}$ is the
	 depth of node $v_{j}$.	 
	 We thus recover the general form of
	 Proposition~\ref{prop:proper_loss}, with
	 \begin{equation*}
		 w_{j} = 
		   \begin{cases}
			   \exp\p{-\alpha d\p{j}} & \text{if $j$ is a leaf node} \\
			   \exp\p{-\alpha d\p{j}}-\exp\p{-\alpha \p{d\p{j}+1}} & \text{otherwise}
		   \end{cases}
		   \enspace,
	 \end{equation*}
	 which leads to a proper scoring rule, as $\displaystyle{\forall k\in\{1,\ldots,K\}, \sum_{j\in a(k)} w_{j} = \exp\p{-\alpha}}$.
	 Hence, from Proposition \ref{prop:proper_loss},
	 $\loss_{\text{HXE}}$ is a proper scoring rule.
\end{proof}

\renewcommand*{\proofname}{Proof of Equation \ref{eq:wasserstein_distance}}
\begin{proof} 
  Since $\mu$ is only supported on leaves and $\nu$ is supported on
  one leaf, we use $f\p{\bx;\btheta}$ instead of $\mu$ and $y$ instead
  of $\nu$ in Equation \eqref{eq:4}.
  Let $\indic_{v_j}(y)$ be the indicator function returning 1 if $y\in v_{j}$ and 0 otherwise, we have:
  \begin{align*}
    W\p{f\p{\bx;\btheta}, y}
      &=\sum_{v_j\in\mathcal V}w_j\abs{\sum_{k\in v_j}f_k\p{\bx;\btheta}-\indic_{v_j}(y)} \\
      &=1-2\sum_{k=1}^K\sum_{j\in a\p{k}}w_j\indic_{v_j}(y)f_k\p{\bx;\btheta} \\ 
      &=1-2\sum_{k=1}^K\sum_{j\in\lca\p{k, y}}w_jf_k\p{\bx;\btheta}\\
      &=\sum_{k=1}^Kf_k\p{\bx;\btheta}\cdot d_{T}\p{y,k},
  \end{align*}
  where we have used the fact that, for $0\leq x\leq 1$, $\abs{x-\indic_y}=\indic_y+\p{1-2\indic_y}x$, and where we changed the order of summations $\sum_{v_j\in\mathcal V}\sum_{k\in v_j}$ to get $\sum_{k=1}^K\sum_{j\in a\p{k}}$.
\end{proof}

\end{document}